\documentclass[10pt,twocolumn,letterpaper]{article}

\usepackage[pagenumbers]{cvpr} 

\usepackage{graphicx} 
\usepackage{multirow}
\usepackage{makecell}
\usepackage{graphicx}
\usepackage[table]{xcolor}
\usepackage{stfloats}
\definecolor{mine}{HTML}{E6F0FF}
\definecolor{light}{HTML}{F2F2F2}

\usepackage{algorithm}
\usepackage{algpseudocode}
\usepackage{amsmath}
\usepackage{amssymb}
\usepackage{bm}
\usepackage{mathtools}
\usepackage{pifont}
\usepackage{titletoc}
\usepackage{marvosym}

\definecolor{cvprblue}{rgb}{0.21,0.49,0.74}
\usepackage[pagebackref,breaklinks,colorlinks,allcolors=cvprblue]{hyperref}

\newcommand{\cmark}{\ding{51}}%
\newcommand{\xmark}{\ding{55}}%
\newcommand{\cmtt}[1]{{\fontfamily{cmtt}\selectfont #1}}
\newcommand{\name}{PlannerRFT}
\newcommand{\boldparagraph}[1]{\vspace{0.1cm}\noindent{\bf #1}}
\newcolumntype{C}[1]{>{\centering\arraybackslash}p{#1}}

\title{\textit{\name}: Reinforcing Diffusion Planners through Closed-Loop and Sample-Efficient Fine-Tuning}

\author{
    Hongchen Li$^{1,2,3}$ \quad
    Tianyu Li$^{2,3}$  \quad
    Jiazhi Yang$^{3}$ \quad 
    Haochen Tian$^{3}$ \quad
    \\
    Caojun Wang$^{1,2,3}$ \quad
    Lei Shi$^{4}$ \quad
    Mingyang Shang$^{5}$ \quad 
    Zengrong Lin$^{5}$ \quad
    \\
    Gaoqiang Wu$^{5}$ \quad
    Zhihui Hao$^{5}$ \quad 
    Xianpeng Lang$^{5}$  \quad 
    Jia Hu$^{1~\textrm{\Letter}}$ \quad
    Hongyang Li$^{3~\textrm{\Letter}}$
    \\[2mm]
    $^1$ Tongji University \quad
    $^2$ Shanghai Innovation Institute \\
    $^3$ OpenDriveLab at The University of Hong Kong \quad
    $^4$ Meituan  \quad
    $^5$ Li Auto Inc. \quad \\
    \\ \vspace{-12pt}
    {
    \hypersetup{urlcolor=Cerulean}
    \href{https://opendrivelab.com/PlannerRFT}
    {\texttt{{https://opendrivelab.com/PlannerRFT}}}
    }
}

\begin{document}

\maketitle

{\let\thefootnote \relax
\footnote{
\hangindent=1.8em
$^{~\textrm{\Letter}}$ Equal co-advising.\\
Primary contact \texttt{lihongchen@tongji.edu.cn}
}}

\begin{abstract}
Diffusion-based planners have emerged as a promising approach for human-like trajectory generation in autonomous driving. Recent works incorporate reinforcement fine-tuning to enhance the robustness of diffusion planners through reward-oriented optimization in a generation–evaluation loop. However, they struggle to generate multi-modal, scenario-adaptive trajectories, hindering the exploitation efficiency of informative rewards during fine-tuning. To resolve this, we propose PlannerRFT, a sample-efficient reinforcement fine-tuning framework for diffusion-based planners. PlannerRFT adopts a dual-branch optimization that simultaneously refines the trajectory distribution and adaptively guides the denoising process toward more promising exploration, without altering the original inference pipeline. To support parallel learning at scale, we develop nuMax, an optimized simulator that achieves 10 times faster rollout compared to native nuPlan. Extensive experiments shows that PlannerRFT yields state-of-the-art performance with distinct behaviors emerging during the learning process. 
\end{abstract}
\section{Introduction} \label{sec:intro}
Diffusion-based planners 
have recently emerged as a powerful probabilistic paradigm for generating human-like and socially compatible driving trajectories in 
dynamic environments~\cite{DiffusionPolicy,DiffusionPlanner,DiffusionDrive}.
Such planners acquire driving skills from large-scale human demonstrations via imitation learning (IL). 
Despite the capability of modeling dexterous behaviors, these methods suffer from distributional shift and objective misalignment, limiting their robustness and reliability in real-world deployment~\cite{lu2023imitation,RAD,Improving,lin2025modelbased}.

\begin{figure}[t]
    \centering
    \includegraphics[width=\linewidth]{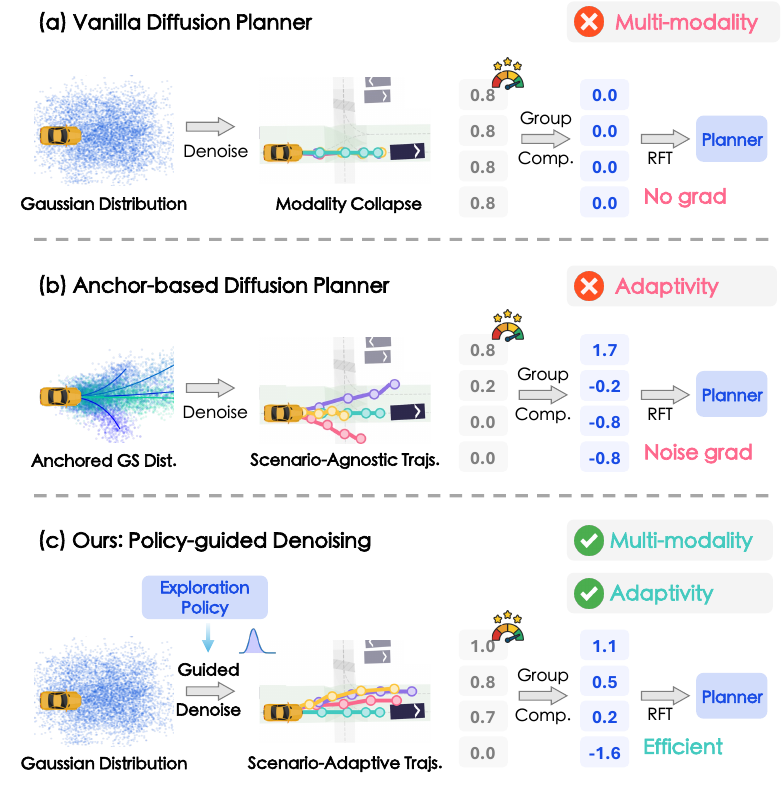}
    \caption{
        \textbf{Comparison to Denoising Strategies} across various
        diffusion planning paradigms.
        \textbf{(a)} Vanilla diffusion planners suffer from mode collapse, offering limited exploration.
        \textbf{(b)} Anchor-based methods
        are oriented towards scenario-agnostic actions, leading to noisy interactions.
        \textbf{(c)} Our policy-guided denoising enables both multi-modal and scenario-adaptive sampling, yielding stable and efficient exploration for optimization.
    }
    \label{fig:teaser}
    \vspace{-9pt}
\end{figure}

Reinforcement learning (RL) offers a potential alternative.
Through simulator-assisted exploration and reward-oriented optimization, RL-based planners can scale with large-scale simulated data and simple rewards~\cite{CaRL,Roach,GIGAFLOW}.
Recent \textit{generation-evaluation} reinforcement fine-tuning (RFT) paradigms~\cite{EvaDrive, GenDrive, DiffusionES, TrajHF} demonstrate the balance between training efficiency and improvements in closed-loop planning performance. In this paradigm, a trajectory generator serves as an actor to produce diverse candidate trajectories, which are then evaluated in simulation and iteratively refined through group-wise reinforcement fine-tuning\cite{shao2024grpo}.
The overall performance of this paradigm primarily depends on the generator’s exploration capability, that is, the distribution of candidate trajectories. This drives two key requirements:
(i) Multi-modality, the ability to generate diverse maneuver hypotheses under the same situation; and
(ii) Adaptivity, the capacity to self-adjust exploration distribution toward more promising behaviors, such as AlphaGo~\cite{AphaGo} used Monte Carlo Tree Search (MCTS)~\cite{MCTS} for adaptive exploration.
However, vanilla diffusion-based planners suffer from modality collapse~\cite{DiffusionDrive}, where trajectories generated from different noise inputs converge to nearly identical results throughout denoising process\footnote{To avoid confusion with the terminology ``trajectory sampling'' in reinforcement learning, we refer to the diffusion sampling process as ``trajectory denoising" in this paper.}, as illustrated in Fig.~\ref{fig:teaser} (a). This collapse limits the exploration capability, leaving reinforcement fine-tuning without an effective optimization signal.
To mitigate this issue, anchor-based diffusion planners \cite{DiffusionDrive, GoalFlow} initialize the denoising process from anchor-centered Gaussian distributions rather than pure Gaussian noise, enabling 
the generation of 
diverse and maneuver-consistent trajectories.
Nevertheless, these fixed, scenario-agnostic anchors are suboptimal for reward-oriented optimization. 
As shown in Fig.~\ref{fig:teaser}(b), a part of anchors yield scene-compatible maneuvers, while many others produce context-conflicting motions, which introduce noisy gradients and hinder stable reinforcement optimization.
Overall, exploration effectiveness requires not only diverse but also scene-consistent maneuvers, which, in turn, facilitate efficient reinforcement fine-tuning.

To this end, we propose \textbf{PlannerRFT}, a closed-loop and sample-efficient framework for diffusion-based \textbf{Planner} \textbf{R}einforcement \textbf{F}ine-tuning.
As shown in Fig.~\ref{fig:teaser} (c), PlannerRFT performs policy-guided denoising to achieve multi-modality and scenario-adaptive trajectory sampling, providing group-wise trajectory optimization with more stable and efficient exploration.
For scalable closed-loop training, we develop a GPU-accelerated simulator, \textbf{nuMax}, which supports high-throughput parallel rollouts.

For multi-modality, PlannerRFT introduces an energy-based classifier guidance~\cite{lu2023contrastive} that injects residual offsets into the denoising process, enabling the model to generate diverse maneuver trajectories.
For adaptivity, a dedicated Exploration Policy learns an adaptive guidance scale to modulate exploration according to scenario context, achieving scenario-aware trajectory generation.
The Exploration Policy is optimized through closed-loop interaction with the simulator using Proximal Policy Optimization (PPO)~\cite{schulman2017ppo}, guiding the planner toward temporally consistent, safe, and efficient behaviors during reinforcement fine-tuning.
For trajectory optimization,  we adopt Group Relative Policy Optimization (GRPO) \cite{shao2024grpo} to fine-tune the diffusion planner denoising process. To stabilize optimization in challenging scenarios, we introduce a survival reward formulation that accumulates non-terminal trajectory rewards, encouraging the planner to delay failure and improve long-horizon viability.
To enhance the scalability and efficiency of online rollouts, we develop \textbf{nuMax}, a GPU-parallel simulator built upon Waymax \cite{Waymax} and calibrated for the large-scale nuPlan benchmark \cite{nuPlan}, achieving up to 10× faster simulation speed than the native nuPlan simulator.
Extensive evaluations on the nuPlan benchmark demonstrate that PlannerRFT achieves state-of-the-art performance.
Compared with the IL-pretrained baseline, PlannerRFT demonstrates notable gains in handling failure scenarios such as collisions and off-road events, leading to improved driving safety.
Furthermore, PlannerRFT exhibits distinct, human-like driving behaviors, with safer and more efficient maneuvers, thereby highlighting the effectiveness of our reinforcement fine-tuning framework.
We summarize our contributions as follows:
\begin{itemize}
    \item We present PlannerRFT, a closed-loop reinforcement fine-tuning framework for diffusion-based planners that enhances the RL sampling efficiency through policy-guided denoising.
    \item We design an exploration policy that adaptively modulates trajectory sampling across scenarios and cooperates with group-wise reinforcement optimization for stable fine-tuning. To support large-scale online training, we further develop nuMax, a GPU-parallel simulator calibrated for the nuPlan benchmark.
    \item Extensive experiments on nuPlan demonstrate that PlannerRFT achieves state-of-the-art performance, while notably enhancing safety and robustness in challenging driving scenarios.
\end{itemize}
\section{Related Work} \label{sec:related work}

\boldparagraph{Diffusion Planners for Autonomous Driving.}
Recently, diffusion models have been widely applied to decision-making and planning tasks in autonomous driving, including motion planning~\cite{DiffusionPlanner,FlowPlanner,FlowDrive}, traffic simulation~\cite{Nexus, Motiondiffuser, Scenediffuser}, and end-to-end driving policy learning~\cite{ReCogDrive, Diffvla, Diffvla++, BridgeDrive, DiffusionDrive}.
Representative works include Diffusion Planner~\cite{DiffusionPlanner}, which jointly models surrounding agents’ trajectories and ego-vehicle planning; Nexus~\cite{Nexus}, which introduces flexible noise scheduling to balance reactivity and goal orientation for the traffic scenario simulation; DiffusionDrive~\cite{DiffusionDrive}, which generates multimodal trajectories via a truncated diffusion process for end-to-end driving; and RecogDrive~\cite{ReCogDrive}, which incorporates vision-language tokens followed by a diffusion head for trajectory generation.
Beyond modeling complex distributions, diffusion planners offer strong flexibility through guidance-based denoising~\cite{DiffusionPlanner,FlowPlanner,FlowDrive} that enables controllable trajectory generation.
However, rule-based guidance strategies introduce competing gradient signals (e.g., between collision avoidance and ride comfort) and impose a fixed guidance strength~\cite{FlowPlanner}. These limitations lead to substantial performance variability across diverse driving scenarios.

\begin{figure*}[t!]
    \centering
    \includegraphics[width=\linewidth]{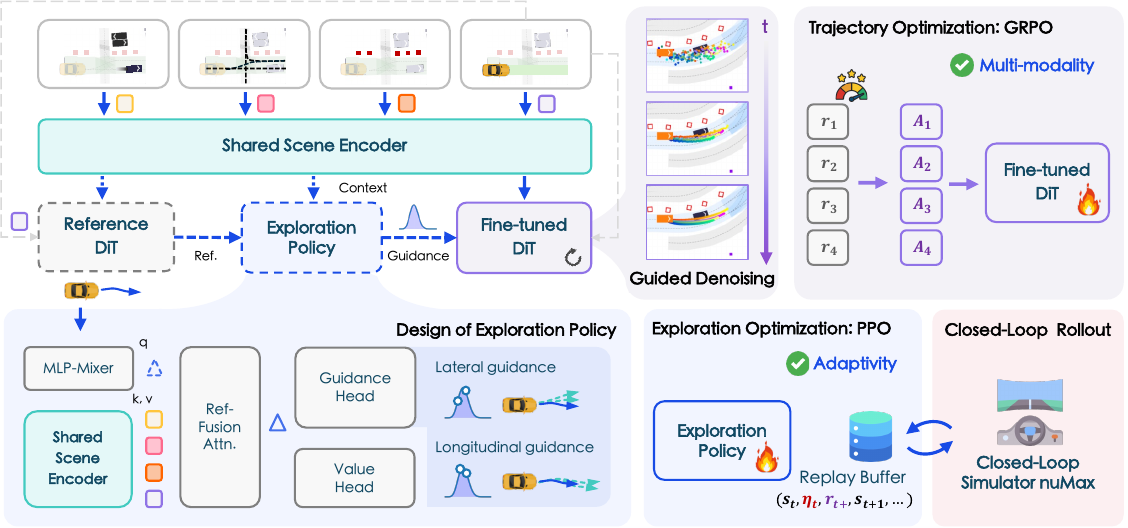}
    \caption{
    \textbf{Overview of \name.} 
    We enhance multi-modality during RL sampling through \textit{Guided Denoising}, with guidance scales modulated by the \textit{Exploration Policy} to generate scenario-adaptive trajectories (\cref{sec:exploration}).
    The planner gathers on-policy interaction data through \textit{Closed-Loop Rollout} in simulation (\cref{sec:rollout}).
    A dual-branch optimization framework performs \textit{Trajectory Optimization} and \textit{Exploration Optimization} to steer the denoising process (\cref{sec:optim}).
    }
    \label{fig:architecture}
\end{figure*}

\boldparagraph{Reinforcement Fine-tuning for Driving Planners.}
Reinforcement learning leverages probabilistic modeling to enable sampling-based exploration and policy optimization. Recent studies on reinforcement fine-tuning for driving planners generally follow three paradigms.
The first discretizes trajectories into a vocabulary of motion tokens~\cite{HydraMDP++,HydraMDP,HydraNext}, optimizing the selection probability of each token under different driving scenarios~\cite{liu2025reinforced, AutoVLA, hamdan2025eta}, similar to RFT in LLMs.
Yet, such discretization inherently constrains planner expressiveness, as a larger token set better captures trajectory diversity but also increases computational complexity and optimization dimensionality.
Another paradigm models each trajectory step as a continuous distribution via auto-regressive generation~\cite{zhang2025carplanner,Plan-R1, Think2Drive}, which inherently suffers from error accumulation and temporal instability across sequential decisions.
Diffusion models exhibit an inherent advantage, as their denoising process operates in a probabilistic manner to generate actions, making them well-suited for reinforcement learning in continuous action spaces and temporally consistent decision processes. However, diffusion-based planners in autonomous driving tend to exhibit modality collapse, which restricts exploration during RFT and consequently hinders effective policy adaptation.

\section{Preliminary}

\boldparagraph{Task Definition.}
Motion planning aims to generate safe and feasible trajectories for the ego vehicle in dynamic driving environments~\cite{liu2024reasoning, zhang2025perception, jia2023driveadapter}.
This work focuses on enhancing the closed-loop performance of IL-pretrained diffusion planners via reinforcement fine-tuning, yielding improved safety, comfort, and efficiency in motion planning.

\boldparagraph{Planner Architecture.}
We adopt a 
pretrained diffusion planner following the commonly used architecture, consisting of a shared scene encoder and a Diffusion Transformer (DiT)~\cite{dit} decoder.
The scene encoder fuses scene inputs, including surrounding agents, map features, and static obstacles, into the environment representation $F_{\text{scene}}$.
The navigation command is encoded as $F_{\text{navi}}$. 
Given the noisy trajectory samples $\mathbf{x}^{k}$ and diffusion timestep $k$, the DiT decoder iteratively denoises the latent samples, conditioned on both the scene and navigation embeddings, mathematically:
\begin{equation}
    \hat{\mathbf{x}}_{0}^{k} = \text{DiT}_{\theta}\left(\text{MLP}(\mathbf{x}^{k}); F_{\text{scene}}; F_{\text{navi}}; t\right).
\end{equation}

\section{Method} \label{sec:method}
In this section, we begin with an overview of our PlannerRFT in Section~\ref{sec:overview}. We then delve into policy-guided denoising in Section~\ref{sec:exploration}, followed by the closed-loop rollout process in Section~\ref{sec:rollout} and the policy optimization in Section~\ref{sec:optim}. Finally, we summarize best practices for PlannerRFT in Section~\ref{sec:bestpractice}.

\subsection{Overview of PlannerRFT} \label{sec:overview}
As illustrated in \cref{fig:architecture}, given an IL-pretrained diffusion planner, PlannerRFT aims to enhance its closed-loop planning performance by adopting the generation–evaluation paradigm with GRPO.
During RFT, the IL-pretrained planner is duplicated and frozen as a global reference.
We introduce policy-guided denoising to improve the multi-modality and adaptivity of the trajectory sampling. To achieve this, we plug in an Exploration Policy on the original model architecture and use closed-loop rollout and PPO to optimize the policy.

\subsection{Policy-guided Denoising}  \label{sec:exploration}

\boldparagraph{Guided Denoising.}  \label{sec:denoising}
Vanilla diffusion planners generate single-pass trajectories and tend to modality collapse, leading to limited trajectory diversity for RL sampling.
To alleviate this limitation and promote exploration, we adopt the energy-based classifier guidance~\cite{dhariwal2021diffusion, lu2023contrastive} that injects residual offsets into the denoising process. 
This enables the planner to generate diverse trajectories in the vicinity of the reference trajectory.
Specifically, we decompose the injected guidance into lateral and longitudinal components.
At each timestep~$\tau$, given the current planner’s predicted waypoints~$\mathbf{x}$ and the reference waypoints~$\mathbf{x}^{\text{ref}}$, the lateral guidance energy function $\Psi_{\text{lat.}}$ is formulated as:
\begin{equation}
    \Psi_{\text{lat.}} = \frac{1}{T}\sum_{\tau=1}^{T} \left(\mathbf{n}_{\tau}^{\perp}\left(\mathbf{x}_{\tau} - \mathbf{x}_{\tau}^{\text{ref}}\right) - \lambda_{\text{lat.}} \eta _{\text{lat.}} \right)^{2},
    \eta _{\text{lat.}}\in [-1, 1],
\end{equation}
where $\mathbf{n}_{\tau}^{\perp}$ is the unit normal vector, $\lambda_{\text{lat.}}$ is the maximum lateral offset (meters), and $\eta _{\text{lat.}}$ is the lateral guidance scale.
The longitudinal guidance modulates the deviation of the planned velocity~$\mathbf{v}$ with the reference velocity~$\mathbf{v}^{\text{ref}}$, as:
\begin{equation}
    \Psi_{\text{lon.}} = \frac{1}{T}\sum_{\tau=1}^{T}
    \left(
    \mathbf{n}_{\tau}^{\parallel}\!\big(\mathbf{v}_{\tau}
    - \lambda_{\text{lon.}}\eta _{\text{lon.}}\mathbf{v}_{\tau}^{\text{ref}}\big)
    \right)^{2},
    \eta _{\text{lon.}}\in[-1,1],
\end{equation}
where $\mathbf{n}_{\tau}^{\parallel}$ is the unit tangent vector, $\lambda_{\text{lon.}}$ is a constant maximum relative speed deviation (percentage), and $\eta _{\text{lon.}}$ is the longitudinal guidance scale. 
These two energy functions yield decoupled and orthogonal gradients, enabling multi-modal trajectory generation through different combinations of $(\eta_{\text{lat.}}, \eta_{\text{lon.}})$.
No explicit map- or vehicle-level collision constraints are imposed; this simplified guidance formulation allows infeasible samples to act as negative feedback for RL optimization.

\boldparagraph{Design of Exploration Policy.}  \label{sec:Exploration}
We introduce the Exploration Policy module, which learns to modulate the guidance scales $(\eta_{\text{lat.}}, \eta_{\text{lon.}})$ conditioned on driving contexts $\mathbf{s}$ and reference waypoints.
This learnable exploration enables the planner to generate context-aware maneuvers, thereby improving exploration effectiveness during RL sampling.
Formulated as:
\begin{equation}
    \boldsymbol{\eta } \sim \pi_{\phi}(\cdot \mid \mathbf{s}, \mathbf{x}^{\text{ref}}).
\end{equation}
Concretely, we use the reference trajectory as a frozen prior to provide PlannerRFT with a stable and well-trained imitation-learning distribution. The reference trajectory is encoded through an MLP-Mixer into a compact token and fused with the scene embedding via a cross-attention module, capturing the interaction between the reference motion and the surrounding environment. Based on this fused representation, the Guidance Head predicts the parameters of two Beta distributions governing the lateral and longitudinal guidance scales. 
In parallel, the Value Head $V_{\psi}$ estimates the state-value $V(s_t)$ to assist policy optimization.

\begin{figure}[t]
    \centering
    \includegraphics[width=\linewidth]{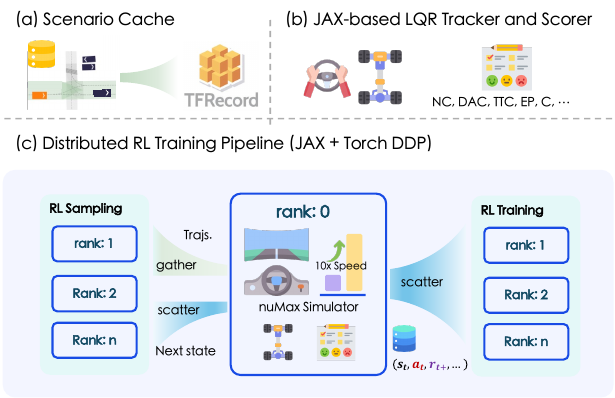}
    \caption{
        \textbf{Illustration of nuMax.} 
        \textbf{(a)} Scenario cache: nuPlan scenarios are preprocessed and cached for fast loading during large-scale RL rollouts; 
        \textbf{(b)} LQR tracker and scorer: vehicle kinematics and reward computation are calibrated to match nuPlan; and 
        \textbf{(c)} Distributed RL training pipeline: enables communication between PyTorch DistributedDataParallel (DDP) workers and the JAX-based simulator.
    }
    \label{fig:simulator}
\end{figure}

\boldparagraph{Trajectory Sampling.}
During RFT, we repeatedly sample the guidance scales 
$(\eta _{\text{lat.}}^{(k)}, \eta _{\text{lon.}}^{(k)})$ 
from the Beta distributions learned by the Exploration Policy. 
Each sampled pair specifies a distinct driving modality and modulates the guided denoising process toward a corresponding trajectory $\hat{\mathbf{x}}^{(k)}$. 
Repeating this process $K$ times yields a diverse set of trajectories
$\mathcal{X} = \{\hat{\mathbf{x}}^{(k)},(\eta _{\text{lat.}}^{(k)}, \eta _{\text{lon.}}^{(k)})\}_{k=1}^{K}$. 
Formally, the Exploration Policy dynamically modulates the classifier-guided denoising gradients as:
\begin{equation} \label{eq:guided}
    \nabla_\mathbf{x} \log p(\mathbf{\eta }|\mathbf{x}) \approx 
    -\nabla_\mathbf{x} \big[\Psi_{\text{lat.}}(\mathbf{x}; \eta _{\text{lat.}}) + \Psi_{\text{lon.}}(\mathbf{x}; \eta _{\text{lon.}})\big],
\end{equation}
thereby enabling the Fine-tuned DiT to produce adaptive and human-like trajectories across scenarios.

\subsection{Closed-loop Rollout} \label{sec:rollout}

\boldparagraph{The nuMax Simulator.} 
Unlike IL methods trained on pre-collected offline datasets, RL is trained on simulated data that is collected during the training process.
Therefore, enhancing simulation throughput is essential for accelerating model iteration and achieving scalable training, given limited computational resources.
To this end, we develop nuMax, a GPU-parallel simulator that enables 10 times faster rollout speed compared with the native nuPlan simulator. 
Our implementation builds upon Waymax~\cite{Waymax} and V-Max~\cite{vmax}, with further implementation details provided in the supplementary material.

\boldparagraph{Rollout Planning.}
At each simulation step, the fine-tuned planner generates a set of $K$ candidate trajectories $\mathbf{\mathcal{X}}$ under different guidance scales. 
To provide diverse training experiences for reinforcement learning, one trajectory $\mathbf{x}^{'}$ and its corresponding guidance scales 
$(\eta _{\text{lat.}}', \eta _{\text{lon.}}')$ 
are randomly selected from the candidate set. 
Only the first action of the selected trajectory is executed in the closed-loop simulator to update the environment state from $s_t$ to $s_{t+1}$ and obtain the immediate reward $r_t$. The current state $s_t$, the selected guidance scales $(\eta _{\text{lat.}}', \eta _{\text{lon.}}')$, and the received reward $r_{t+1}$ are stored in the replay buffer $\mathcal{B}$ for subsequent policy updates $(s_{t}, \eta _{\text{lat.}}^{'}, \eta _{\text{lon.}}', r_{t+1}, V(s_{t}))$. 

\subsection{Policy Optimization} \label{sec:optim}

\boldparagraph{Exploration Policy Optimization.}
The Exploration Policy $\pi_{\phi}$  is optimized following the PPO framework. Specifically, the goal of this optimization is to provide temporally consistent, efficient, safe, and comfortable exploration directions during closed-loop planning.
Future rewards are propagated backward through Generalized Advantage Estimation (GAE), allowing the policy to refine its current exploratory decisions based on the long-term trajectory performance observed in closed-loop rollouts.
Through iterative rollouts and updates, the Exploration Policy learns Beta-distribution parameters that adaptively set $(\eta_{\text{lat}}, \eta_{\text{lon}})$ to the driving context, improving exploration effectiveness.

\boldparagraph{Trajectory Optimization.}
The Fine-tuned DiT focuses on long-horizon planning conditioned on the current scenario.
We evaluate the trajectory based on the Predictive Driver Model Score (PDMS) over a prediction horizon $T_{r}$ in an open-loop manner.
However, direct use of the terminal reward (collision and off-road) leads to optimization stagnation in hard scenarios, as all candidate trajectories collapse to zero reward once a failure occurs, resulting in no optimization gradient within the group.
To alleviate this issue, we introduce a \textit{survival reward} formulation that accumulates trajectory-level rewards only over valid, non-terminal segments.
Formally, given a per-step termination reward sequence ${R^{\text{term}}}_{\tau=1}^{T_{r}}$, the survival reward is defined as:
\begin{equation}
R_{\text{surv}} = \frac{1}{T_{r}}\sum_{\tau=1}^{T_{r}} R_{\tau}^{\text{term}}
\prod_{j=1}^{\tau} \mathbb{I}[R_{j}^{\text{term}} \neq 0].
\label{eq:survival}
\end{equation}
This formulation encourages the planner to optimize toward trajectories that delay the failure event, improving exploration in hard scenarios.

We fine-tune the planner’s trajectory distribution via the GRPO framework. Following DPPO~\cite{DPPO} and ReCogDrive~\cite{ReCogDrive}, the diffusion denoising process is formulated as a Markov Decision Process, where each denoising step represents a Gaussian transition. By updating the Gaussian parameters during RFT, the planner better aligns with reward-oriented objectives, improving closed-loop stability and planning performance.

\subsection{Best Practices for PlannerRFT} \label{sec:bestpractice}

We summarize the best practices for effectively fine-tuning diffusion-based planners with PlannerRFT as follows.

\boldparagraph{Fine-tune DDIM Denoising.}
We adopt a 5-step DDIM~\cite{ddim} denoising scheme.
Compared with ODE-based denoising, DDIM introduces stochasticity that enhances exploration, while requiring far fewer steps than DDPM~\cite{DDPM} to maintain high training efficiency.

\boldparagraph{Zero-initialization of Exploration Policy.}
The Exploration Policy is initialized to produce zero-mean lateral and longitudinal guidance scales.
This initialization ensures unbiased exploration around the reference trajectory and mitigates performance drops at the early stage of fine-tuning.

\boldparagraph{Plug-and-play Fine-tuning.}
During RFT, the Reference DiT and Exploration Policy are integrated to guide the denoising process, facilitating exploration and policy refinement.
At deployment, these modules are removed, enabling the planner to retain its original diffusion structure while delivering improved trajectory performance.

\boldparagraph{Hard-case Fine-tuning.}
Incorporating a moderate proportion of challenging scenarios significantly improves the planner’s robustness, while an excessively hard training set may degrade overall performance. Further analysis of fine-tuning data selection is provided in the Section~\ref{sec:ablation}.

\begin{figure*}[t]
    \centering
    \includegraphics[width=\linewidth]{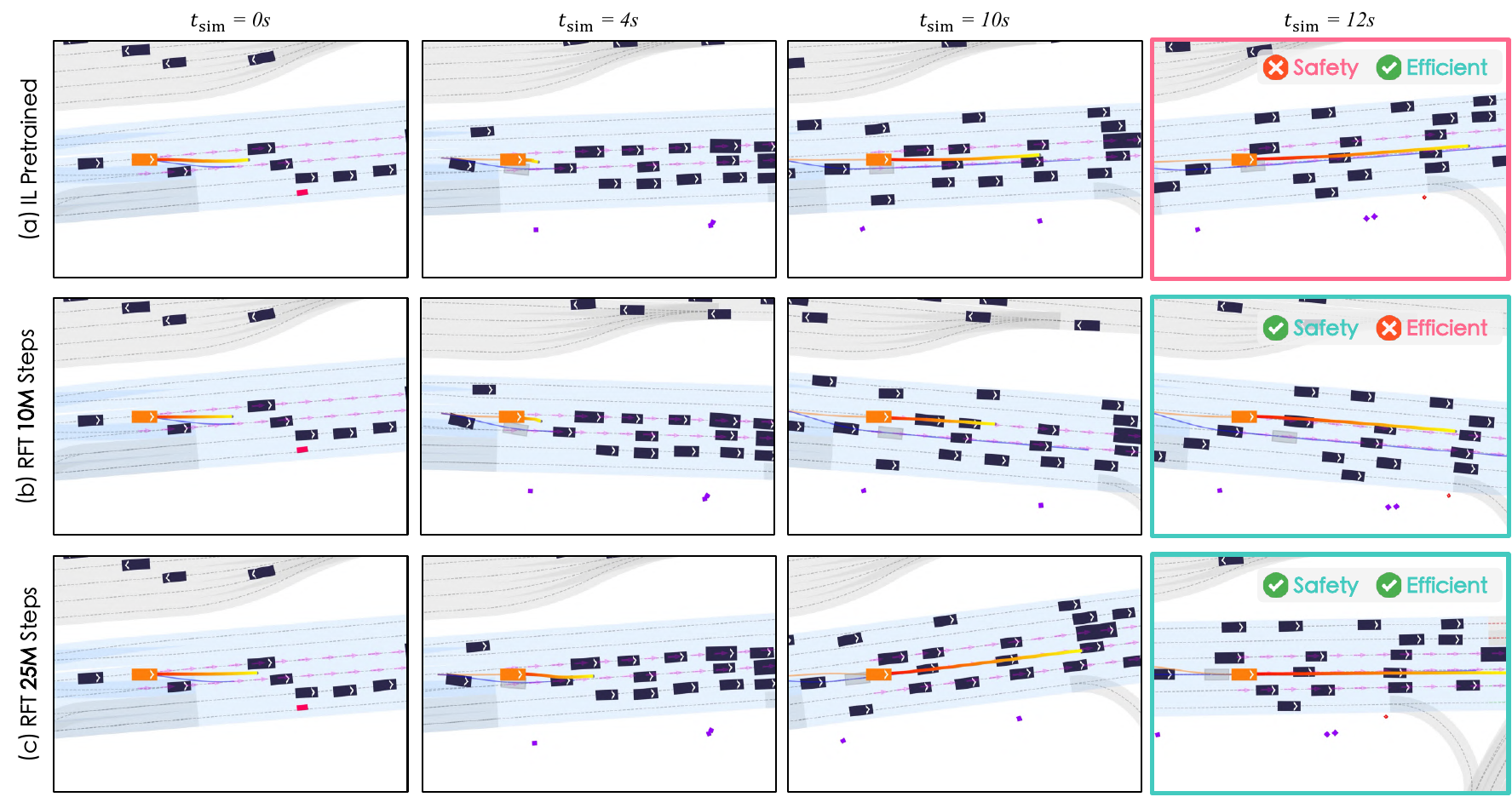}
    \vspace{-18pt}
    \caption{
        \textbf{Qualitative Comparison of Pretrained Planner and RFT Planner.}
        In each frame shot, the \textcolor{orange}{simulation position} and \textcolor{orange}{planning trajectory} are marked as \textcolor{orange}{orange}, the \textcolor{gray}{ground-truth position} and \textcolor{blue}{ground-truth trajectory} recorded in the driving log are marked as \textcolor{gray}{gray} and \textcolor{blue}{blue}, respectively.
        }
    \label{fig:demo_scenario}
    \vspace{-12pt}
\end{figure*}

\section{Experiments}

This section aims to explore the following research questions:  
\textbf{1)} Can PlannerRFT improve the closed-loop planning performance of diffusion-based planners through reinforcement fine-tuning?  
\textbf{2)} Does the Exploration Policy enhance sample efficiency through the policy guided denoising?
\textbf{3)} Will the fine-tuned planner exhibit distinct behavioral patterns from imitation learning?
\textbf{4)} What are the key factors that influence the effectiveness of RFT training?

\subsection{Setup and Protocals}

\boldparagraph{Benchmarks and Baselines.}
We evaluate PlannerRFT on the large-scale nuPlan benchmark~\cite{nuPlan}.
The Val14 benchmark~\cite{Val14} is used to assess model performance under general driving scenarios, while the Test14-hard benchmark~\cite{Test14H} includes more complex and challenging situations, reflecting the model's robustness in hardcore scenarios.
All evaluations are performed within the nuPlan closed-loop simulator, which supports both non-reactive and reactive background traffic settings. In the non-reactive setting, surrounding vehicles follow pre-recorded trajectories; in contrast, the reactive setting employs an Intelligent Driver Model (IDM)~\cite{IDM} that dynamically adjusts surrounding vehicles' behaviors according to the ego vehicle's actions, providing a more realistic simulation of real-world interactions. We compare PlannerRFT against a wide range of baseline methods, including rule-based planners (IDM~\cite{IDM}, PDM-Closed~\cite{Val14}), learning-based planners (PlanTF~\cite{Test14H}, GameFormer~\cite{Gameformer}, PLUTO~\cite{Pluto}), and recent generative planning approaches (Diffusion Planner~\cite{DiffusionPlanner}, Flow Planner~\cite{FlowPlanner}). 
The final evaluation score is computed as the average across all scenarios, ranging from 0 to 100, where a higher score indicates better planning performance.

\begin{table}[t]
\centering
\caption{
    \textbf{Closed-loop Planning Results on nuPlan Dataset.}
    The \textbf{highest} and the \underline{second-best} results of each benchmark are denoted by \textbf{bold} and \underline{underline}.
}
    \vspace{-3pt}
\scalebox{0.8}{
    {\setlength{\tabcolsep}{5pt}
    \begin{tabular}{@{}llccccccccc@{}}
    \toprule
    \multirow{2}{*}[-0.15ex]{\makecell[l]{\textbf{Type}}}     & \multirow{2}{*}[-0.15ex]{\makecell[l]{\textbf{Planner}}}       & \multicolumn{2}{c}{\textbf{Val14}} & \multicolumn{2}{c}{\textbf{Test14-hard}} \\ 
    \cmidrule(lr){3-4} \cmidrule(lr){5-6}
    &  & \textbf{NR} & \textbf{R} & \textbf{NR} & \textbf{R}\\ \midrule
    \textcolor{gray}{Expert} & \textcolor{gray}{Log-replay} & \textcolor{gray}{93.53} & \textcolor{gray}{80.32} & \textcolor{gray}{85.96} & \textcolor{gray}{68.80} \\ \midrule
    \multirow{2}{*}[-0.15ex]{\makecell[l]{Rule}}
    & IDM         & 75.60  & 77.33  & 56.15  & 62.26  \\ 
    & PDM-Closed  & 92.84  & 92.12  & 65.08  & 75.19  \\
    \midrule
    \multirow{7}{*}[0.5ex]{\makecell[l]{Learning }}
    & PDM-Open  & 53.53  & 54.24  & 33.51  & 35.83  \\  
    & GameFormer          & 13.32  & 8.69   & 7.08   & 6.69   \\ 
    & PlanTF              & 84.27  & 76.95  & 69.70  & 61.61  \\ 
    & PLUTO               & 88.89  & 78.11  & 70.03  & 59.74  \\ 
    & Diffusion Planner   & \colorbox{light}{89.87}  & \colorbox{light}{82.80}  & \colorbox{light}{75.99}  & \colorbox{light}{69.22}  \\
    & Flow Planner        & \colorbox{mine}{\textbf{90.43}}  & \underline{83.31}  & \underline{76.47} & \underline{70.42}  \\
    & \textbf{PlannerRFT(Ours)}  & \underline{89.96}  & \colorbox{mine}{\textbf{84.46}}  & \colorbox{mine}{\textbf{77.16}} & \colorbox{mine}{\textbf{72.21}}  \\
    \bottomrule
    \end{tabular}
}}
\label{tab:nuplan}
    \vspace{-15pt}
\end{table}

\boldparagraph{Pretrain.}
We adopt the Diffusion Planner~\cite{DiffusionPlanner} as our IL-pretrained planner, which is trained on 1 million clips from the nuPlan dataset.
We replace the ODE-based DPM-solver~\cite{lu2022dpm} denoising with a 5-step DDIM sampler. Compared with the ODE sampler, the DDIM sampler achieves nearly the same performance while introducing stochasticity that enhances exploration, and its reduced number of denoising steps further improves the efficiency of RL training.

\boldparagraph{Fine-tune Dataset.}
For reinforcement fine-tuning, we collect 144,494 non-overlapping scenarios from nuPlan at 10\,Hz sampling rate.
Each scenario contains 20 frames of history, one current frame, and 150 frames of future trajectory, totaling 171 frames.
We evaluate all scenes using the pretrained planner and construct three datasets according to performance scores:
(1) \cmtt{Fail}, including 10,417 collision or off-road cases;
(2) \cmtt{Lt90}, including all low-score (less than 90) cases, totaling 24,691 scenes;
and (3) \cmtt{All}, which includes all available scenes. 

\boldparagraph{RFT Details.}
All experiments are conducted on 8 NVIDIA H100 GPUs. The fine-tuning process runs for 40M environment steps. Hyperparameters for PPO and GRPO optimization are provided in the supplementary material.

\begin{table*}[htbp!]
\caption{
    \textbf{Ablation on Exploration Policy.}
    $\text{IL Pretrain}_\texttt{DDIM}$ denotes the pretrained Diffusion Planner with 5 steps of DDIM denoising. All planners use the same 5-step DDIM denoising setup.
    $\mathcal{D}$ denotes the modality of the sampled trajectory group, consistent with the definition in DiffusionDrive~\cite{DiffusionDrive}, $\bar{r}$ and $s_{r}$ denote the mean and standard deviation of the corresponding rewards, respectively.
    }
\vspace{-6pt}
\centering
\scalebox{0.8}{
    \begin{tabular}{l|lcccccc|l|c|cc}
        \toprule
        Exploration Type & \textbf{R-score}$\uparrow$ & Collisions & TTC & Drivable & Comfort & Progress & Speed & \textbf{NR-score}$\uparrow$ & $\mathcal{D}~(\%)$ & $\bar{r}\uparrow$& $s_{r}$ \\
        \midrule
        $\text{IL Pretrain}_{\texttt{DDIM}}$ & 68.18 & 86.58 & 79.05 & 94.48 & \underline{86.03} & 76.99 & 97.20 & 76.01 & - & - & - \\
        \midrule
        w/o Guidance & 68.83{\scriptsize \textcolor{blue}{(+0.65)}} & 86.03 & 79.41 & 94.48 & \textbf{87.87} & 77.12 & 97.35 & 76.34{\scriptsize \textcolor{blue}{(+0.33)}} & 5.65 & 69.06 & 0.02 \\
        w/ Uniform Dist. & 65.82{\scriptsize \textcolor{red}{(-2.36)}} & 84.37 & 75.74 & 93.01 & 80.88 & 76.19 & 97.59 & 75.19{\scriptsize \textcolor{red}{(-0.82)}} & 39.78 & 60.44 & 0.12 \\
        w/ Fixed Beta Dist. & \underline{70.65}{\scriptsize \textcolor{blue}{(+2.47)}} & \underline{87.68} & \underline{80.88} & \underline{94.85} & 84.56 & \textbf{77.34} & \underline{97.71} & \underline{76.61}{\scriptsize \textcolor{blue}{(+0.60)}} & 27.73 & \underline{71.50} & 0.07 \\
        \midrule
        \textbf{PlannerRFT(Ours)} & \textbf{72.21{\scriptsize \textcolor{blue}{(+4.03)}}} & \textbf{88.97}{\scriptsize \textcolor{blue}{(+2.39)}} & \textbf{84.93}{\scriptsize \textcolor{blue}{(+5.34)}} & \textbf{95.59}{\scriptsize \textcolor{blue}{(+1.11)}} & 85.66 & \underline{77.17} & \textbf{98.03} & \textbf{77.16{\scriptsize \textcolor{blue}{(+1.15)}}} & 25.34 & \textbf{73.88} & 0.06 \\
        \bottomrule
    \end{tabular}
}
    \label{tab:abl_explorer}
    \vspace{-6pt}
\end{table*}

\subsection{Main Results}

\boldparagraph{Comparison with SOTAs.}
\cref{tab:nuplan} presents the planning results under both challenging (Test14-hard) and general (Val14) test settings.
Compared with the pretrained Diffusion Planner, our PlannerRFT improves closed-loop planning performance across all four benchmarks.
Notably, in reactive traffic settings, PlannerRFT achieves substantial gains, with improvements of $+1.66$ points on the Val14 benchmark and $+2.99$ points on the Test14-hard benchmark.
This suggests that closed-loop rollouts expose the planner to a broader range of interaction patterns, mitigating distribution shift, while the iterative feedback during rollouts enables the model to continuously refine and improve its trajectories.
Compared with other SOTA planners, PlannerRFT achieves the best overall performance in three out of four benchmarks. However, in non-reactive regular scenarios (Val14-NR), the performance improvement remains marginal. 
This may stem from the inherent distributional bias of non-reactive environments.
Notably, PlannerRFT yields a $+2.99$ points improvement on the Test14-hard-NR set, which contains dynamic, interaction-heavy scenarios, highlighting its effectiveness.

\boldparagraph{Qualitative Results.}
Distinct behaviors compared with IL pretraining emerge during reinforcement fine-tuning. Through reward-oriented optimization, the planner adapts its driving policy toward safer and more efficient maneuvers.
\cref{fig:demo_scenario} illustrates an out-of-distribution (OOD) scenario for the IL-pretrained planner.
As shown in \cref{fig:demo_scenario}~(a), the pretrained planner attempts a lane change but fails to handle interactive conflicts, causing the ego vehicle to get stuck between two lanes and collide at $t_{\text{sim}}=12\text{s}$.
After 10M fine-tuning steps, as shown in \cref{fig:demo_scenario}~(b), the planner learns to avoid the collision through lane keeping, while achieving safety, but at the cost of efficiency.
With 25M steps, as shown in \cref{fig:demo_scenario}~(c), the planner executes a decisive lane-change maneuver, achieving both safety and efficiency.

\begin{figure}[t]
    \centering
    \includegraphics[width=\linewidth]{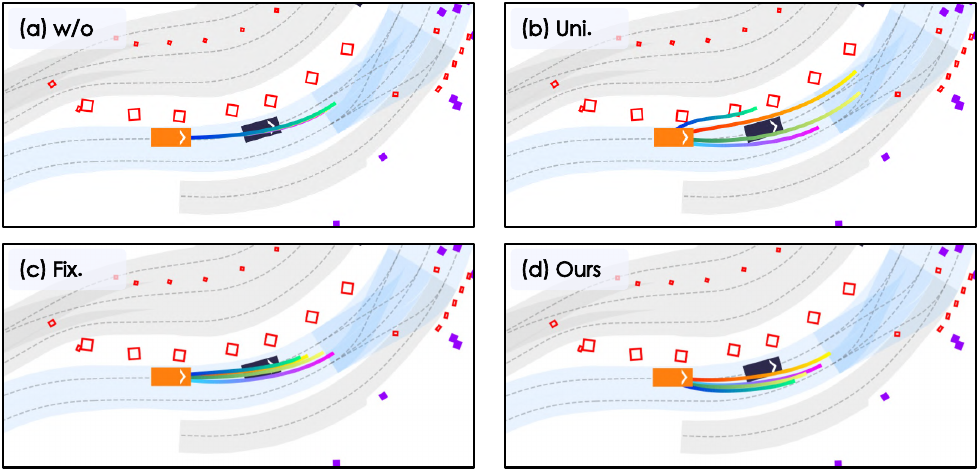}
    \vspace{-15pt}
    \caption{
        \textbf{Visualization of Different Exploration Policies.}
        \textbf{(a)} Without guidance: denoising from random noise.
        \textbf{(b)} Uniform exploration policy: $(\eta_{\text{lat.}}, \eta_{\text{lon.}})$ are sampled from a uniform distribution.
        \textbf{(c)} Fixed exploration policy: $(\eta_{\text{lat.}}, \eta_{\text{lon.}})$ are sampled from the non-learnable Beta distribution initialized from the Exploration Policy’s zero parameters.
        \textbf{(d)} Our policy-guided denoising: exploration adapt to the driving context.
    }
    \label{fig:sampling}
    \vspace{-12pt}
\end{figure}

\subsection{Ablation Study} \label{sec:ablation}

\boldparagraph{Effectiveness of Exploration Policy.}
We evaluate four exploration policies in RL sampling:  
1) denoising from random noise without any guidance,  
2) sampling the guidance scale from a uniform distribution, $\mathbf{\eta} \sim \mathcal{U}(-\mathbf{\lambda}, \mathbf{\lambda})$, 
3) sampling the guidance scale from a fixed Beta distribution,
and 4) our policy-guided denoising.

As for multi-modality, we use the diversity score $\mathcal{D}$ from the DiffusionDrive
as the evaluation metric, which is based on the mean Intersection over Union (mIoU) between each sampled trajectory and the others of the trajectory group. A higher diversity score $\mathcal{D}$ means less trajectory overlap, in terms of more exploration variety. As shown in \cref{tab:abl_explorer}, compared with denoising from random noise, our guided denoising improves trajectory diversity during RL sampling. 
The visualization results in \cref{fig:sampling} demonstrate that our guidance enables the planner to generate continuous and smooth trajectories around the reference trajectory, enhancing both lateral and longitudinal diversity, thereby enhancing exploration.

\begin{figure*}[t]
    \centering
    \includegraphics[width=\linewidth]{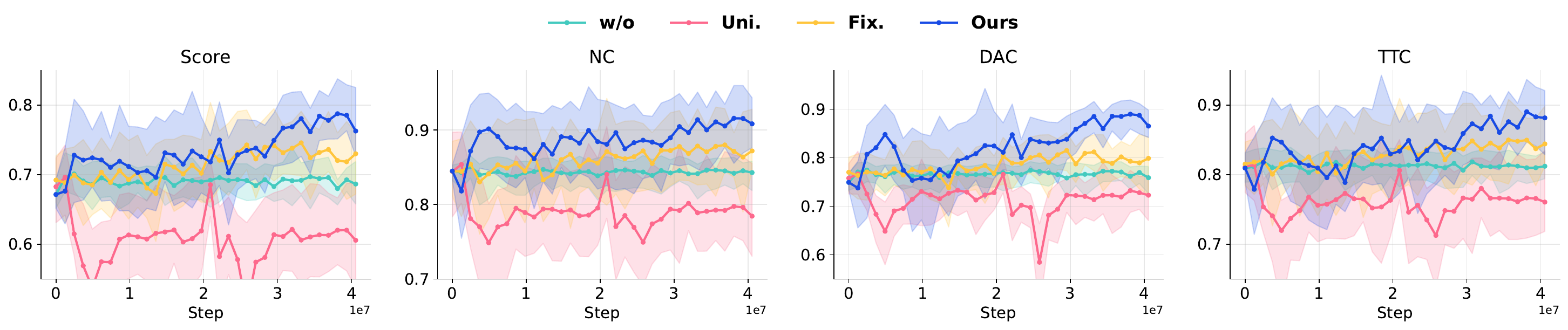}
    \vspace{-15pt}
    \caption{
    \textbf{Closed-loop performance of safety metrics during training under different exploration policy.}
    Score denotes the nuPlan aggregate score, NC refers to No at-fault Collisions, DAC represents Drivable Area Compliance, and TTC indicates Time-to-Collision. Our adaptive exploration policy achieves consistently higher performance and stability across all metrics compared to fixed, uniform, and unguided exploration baselines.
    }
    \label{fig:training_logs}
    \vspace{-6pt}
\end{figure*}

For adaptivity, we compute the mean and standard deviation of rewards across all trajectories sampled in each GRPO group. As shown in \cref{tab:abl_explorer}, the uniform exploration policy yields the highest diversity score but also the worst performance. This is because its scenario-agnostic sampling introduces excessively large reward variance, causing training instability and repeated episodes of reward collapse, as illustrated in \cref{fig:training_logs}.
In contrast, the fixed exploration policy stabilizes training by restricting the exploration range, but the overly limited search space also constrains the achievable performance ceiling.
Our policy-guided denoising exploration adaptively adjusts the exploration direction based on the context, achieving both stable training and higher closed-loop performance.

\begin{table*}[htbp]
\centering
\begin{minipage}[t]{0.38\textwidth}
\centering
\caption{
\textbf{Ablations of fine-tuning data distribution.}
}
\vspace{-6pt}
\scalebox{0.7}{
    {\setlength{\tabcolsep}{5pt}
    \begin{tabular}{@{}llccccccccc@{}}
    \toprule
    \multirow{2}{*}[-0.15ex]{\makecell[l]{\textbf{Training} \textbf{Type}}}     & \multirow{2}{*}[-0.15ex]{\makecell[l]{\textbf{Dataset}}}       & \multicolumn{2}{c}{\textbf{Val14}} & \multicolumn{2}{c}{\textbf{Test14-hard}} \\ 
    \cmidrule(lr){3-4} \cmidrule(lr){5-6}
    &  & \textbf{NR} & \textbf{R} & \textbf{NR} & \textbf{R}\\
    \midrule
    \multirow{1}{*}[-0.15ex]{\makecell[l]{IL Pretrain}}
    & \cmtt{All}  & \colorbox{light}{89.87}  & \colorbox{light}{82.80}  & \colorbox{light}{75.99}  & \colorbox{light}{69.22}  \\ 
    \midrule
    \multirow{1}{*}[-0.15ex]{\makecell[l]{IL Fine-tune}}
    & \cmtt{Lt90} & 88.91  & 82.08  & 74.32  & 67.55  \\ 
    \midrule
    \multirow{3}{*}[-0.15ex]{\makecell[l]{RL Fine-tune}}
    & \cmtt{Fail}  & 82.97  & 77.48  & 69.26  & 63.75  \\ 
    & \cmtt{All}  & \underline{89.93}  & \textbf{84.88}  & 75.50  & \underline{70.43}  \\
    & \cmtt{Lt90} & \textbf{89.96}  & \underline{84.46}  & \textbf{77.16}  & \textbf{72.21}  \\
    \bottomrule
    \end{tabular}
}}
    \label{tab:dataset}
\end{minipage}
\hspace{0.005\textwidth}
\begin{minipage}[t]{0.34\textwidth}
\centering
\caption{\textbf{Ablations of the GRPO reward type and reward horizon.}}
\vspace{-4pt}
\scalebox{0.7}{
    {\setlength{\tabcolsep}{5pt}
    \begin{tabular}{llccccccccc}
    \toprule
    \multirow{2}{*}{\makecell[l]{\textbf{Reward}\\ \textbf{Type}}}     & \multirow{2}{*}[-0.15ex]{\makecell[l]{\textbf{Horizon}\\ \textbf{(s)}}}       & \multicolumn{2}{c}{\textbf{Val14}} & \multicolumn{2}{c}{\textbf{Test14-hard}} \\ 
    \cmidrule(lr){3-4} \cmidrule(lr){5-6}
    &  & \textbf{NR} & \textbf{R} & \textbf{NR} & \textbf{R}\\
    \midrule
    \multirow{1}{*}{\makecell[l]{Terminal}}
    & 4  & \underline{89.78}  & 84.27  & 76.81  & 71.59 \\ 
    \midrule
    \multirow{3}{*}{\makecell[l]{Survival}}
    & 2  & 89.54  & 84.08  & 76.49  & 70.10  \\ 
    & 4  & \textbf{89.96}  & \textbf{84.46}  & \textbf{77.16}  & \textbf{72.21}  \\
    & 6  & 89.66  & \underline{84.31}  & \underline{76.96}  & \underline{71.91}  \\
    \bottomrule
    \end{tabular}
}}
\label{tab:reward}
\end{minipage}
\hspace{0.005\textwidth}
\begin{minipage}[t]{0.25\textwidth}
\centering
\caption{\textbf{Ablations of the maximum guidance offset $\lambda$} on the Test14-hard Reactive benchmark.}
\vspace{-4pt}
\scalebox{0.7}{
    \setlength{\tabcolsep}{5pt}
    \begin{tabular}{lccc}
    \toprule
    & \multicolumn{3}{c}{$\lambda_{\text{lon.}}$ (\%)} \\ 
    \cmidrule(lr){2-4}
    $\lambda_{\text{lat.}~(m)}$  & \textbf{10} & \textbf{25} & \textbf{50} \\
    \midrule
    \textbf{1.0} & 69.94 & 71.41 & 70.26 \\
    \textbf{2.5} & 70.64 & \textbf{72.21} & 71.95 \\
    \textbf{5.0} & 70.11 & 71.63 & 69.99 \\
    \bottomrule
    \end{tabular}
}
\label{tab:lambda}
\end{minipage}
\end{table*}

\boldparagraph{Impact of Fine-tuning Data Distribution.}
We find that the composition of training scenarios substantially alters the characteristics of the learning process.
As shown in Table~\ref{tab:dataset}, training exclusively on collision cases (\cmtt{Fail}) causes severe performance degradation across all benchmarks, indicating that overly hard scenarios can make the planner forget how to handle regular driving maneuvers.
In contrast, training on all available scenarios (\cmtt{All}) includes a large number of easy cases, leading to a weak optimization signal and limited gains on hard scenarios.
The best results are obtained when fine-tuning on a balanced dataset (\cmtt{Lt90}) that combines collision and low-score cases. This suggests that an appropriate proportion of hard cases is essential for effective RFT.
For completeness, we also include an IL fine-tuning baseline trained on the same \cmtt{Lt90} dataset. The IL-finetuned model performs worse, confirming that PlannerRFT’s gains arise from effectively learning under the hard training distribution through exploration, rather than from additional training iterations.

\boldparagraph{Effect of Reward Type and Horizon.}
Table~\ref{tab:reward} compares different reward formulations and horizons of the GRPO reward.
The terminal reward performs comparably to survival on Val14 but degrades on Test14-hard, where collisions or off-route events frequently reset the reward to zero.
In contrast, the survival reward encourages trajectories that delay failure, enabling continuous improvement in closed-loop settings.
For the reward horizon, a short 2 $s$ horizon underperforms due to limited temporal context, while 4 $s$ and 6 $s$ horizons yield similar results, suggesting that a moderate horizon length is sufficient for fine-tuning.

\boldparagraph{Effect of the Maximum Guidance Offset $\lambda$.}
We grid search the maximum lateral and longitudinal guidance offset $\lambda$, as shown in \cref{tab:lambda}.
A small $\lambda$ limits exploration and constrains policy optimization, while a large $\lambda$ drives the policy too far away from the human expert behavior distribution. Both 
challenge the optimization stability.
Instead, a moderate $\lambda$ thus provides an appropriate trade-off between exploration and exploitation.
\section{Conclusion and Outlook}

In this paper, we present PlannerRFT, a closed-loop and sample-efficient reinforcement fine-tuning framework for diffusion-based planners. 
Experiments on nuPlan verify its improvements in closed-loop performance. 
Comparisons with an IL fine-tuning baseline show that these gains arise from effective exploration rather than additional training iterations. 
Analyses of different exploration policies further highlight PlannerRFT’s scenario-adaptive advantage in sample efficiency.

\boldparagraph{Limitations and Future Work.}
PlannerRFT is currently verified on planners with 
structured abstract inputs, instead of sensory observations like images.
Its applicability to visuomotor planners remains underexplored~\cite{cao2025pseudo, jiang2025realengine, li2025mtgs}.
Nonetheless, its sample-efficient designs upon a pretrained policy
laid the foundation of training end-to-end planners in a closed-loop manner with RL, which is left as our future work.

\section*{Acknowledgments}
This work is in part supported by 
the Project supported by the Young Scientists Fund of the National Natural Science Foundation of China (Grant No. 62206172), 
National Natural Science Foundation of China (Grant No. 52372317)
and 
the Beijing Nova Program.
We also appreciate the general research sponsorship from Li Auto.

{
    \small
    \bibliographystyle{ieeenat_fullname}
    \bibliography{main}
}

\clearpage
\appendix
\maketitlesupplementary

\setcounter{page}{1}
\renewcommand{\thefigure}{A\arabic{figure}}
\renewcommand{\thetable}{A\arabic{table}}
\renewcommand{\theequation}{A\arabic{equation}}
\setcounter{figure}{0}
\setcounter{table}{0}
\setcounter{equation}{0}

\section{Discussions}
Towards a better understanding of this work, we supplement intuitive questions that may raise. 

\boldparagraph{Q1.} \textit{What makes PlannerRFT effective?}
PlannerRFT’s effectiveness stems from three key factors:

\textbf{Enhance Lateral movement.}
The expert trajectory distribution is dominated by straight-driving maneuvers, causing IL planners to underfit lateral skills such as lane changes and obstacle avoidance.
PlannerRFT introduces structured lateral perturbations through guided denoising and reinforces them with reward-oriented optimization, enabling lane changes and obstacle-avoidance behaviors in complex scenes, as shown in \cref{fig:demo_obstacle1} and \cref{fig:demo_scenario}.

\textbf{Plan-Motion Alignment.}
IL planners mimic expert trajectories without considering the downstream controller, leading to execution failures in narrow or high-precision maneuvers.
PlannerRFT evaluates executed closed-loop trajectories via the simulator’s vehicle dynamics and updates the planner with execution-level rewards such as collision and off-road penalties.
This closed-loop correction bridges the gap between planning and execution and markedly improves maneuver precision and controller feasibility in challenging environments, as shown in~\cref{fig:demo_obstacle2}.

\textbf{Trial-and-error Rollouts.}
Real traffic is dynamic, and blindly reproducing expert behavior can be unsafe.
PlannerRFT leverages trial-and-error closed-loop RL to adapt to dynamic traffic, markedly improving interaction capability, as shown in \cref{fig:demo_reactive1} and \cref{fig:demo_reactive2}, which also explains its strong gains on reactive-traffic benchmarks.
We further observe that RL helps mitigate the causal-confusion issues inherent in IL, as shown in \cref{fig:demo_confusion}.

\boldparagraph{Q2.} \textit{Why adopt a dual-branch optimization (PPO and GRPO), and how is training stability maintained?}
The exploration policy adjusts guidance scales at every simulation step and directly affects long-horizon closed-loop behavior, making PPO suitable for online learning. In contrast, the DiT generator outputs high-dimensional multi-step trajectories that are better optimized offline with group-based updates, where GRPO provides efficient training. Training stability is further ensured by applying policy-guided denoising around a fixed reference trajectory rather than the evolving DiT outputs, and the well-trained reference trajectory distribution prevents collapse and yields steadily improving rewards, as shown in \cref{fig:training_logs}.

\begin{algorithm}[t]
    \caption{\textbf{Guided Denoising for RL Sampling}}
    \label{alg:rl_sampling}
    \begin{algorithmic}[1]
    \Require 
    Current observation $o_{t}$,
    scene encoder $E_{\text{scene}}$, 
    route encoder $E_{\text{navi}}$,
    reference DiT $D_{\text{ref}}$,
    fine-tuned DiT $D_{\theta}$,
    exploration policy $\pi_{\phi}$,
    GRPO group size $G_{\text{grpo}}$
    
    \State \textcolor{gray}{\textbf{// Step 1: Scenario encoding.}}
    \State $F_{\text{scene}} \gets E_{\text{scene}}(o_t)$
    \State $F_{\text{navi}} \gets E_{\text{navi}}(o_t)$
    
    \State \textcolor{gray}{\textbf{// Step 2: Get reference trajectory.}}
    \State $x^{\text{ref}}_{S} \gets z$, $z \sim \mathcal{N}(\mathbf{0}, \mathbf{I})$
    \For{$i = 1$ to $S$}   \Comment{$S=5$ DDIM steps}
        \State $s \gets s_i$  \Comment{VP-SDE timestep}
        \State $x^{\text{ref}}_{s-1} \gets 
            D_{\text{ref}}(x^{\text{ref}}_{s}, s, F_{\text{scene}}, F_{\text{navi}})$
    \EndFor
    \State $x^{\text{ref}} \gets x^{\text{ref}}_{0}$
    
    \State \textcolor{gray}{\textbf{// Step 3: Get adaptive exploration direction.}}
    \State $(a_{\text{lat.}}, b_{\text{lat.}}, a_{\text{lon.}}, b_{\text{lon}}) \gets 
           \pi_{\phi}\!\left(x^{\text{ref}}, F_{\text{scene}}, F_{\text{navi}}\right)$
    
    \State \textcolor{gray}{\textbf{// Step 4: Sample multi-modal guidance scales.}}
    \For{$k = 1$ to $G_{\text{grpo}}$}
        \State $\eta_{\text{lat}}^{(k)} \sim \mathrm{Beta}(a_{\text{lat}}, b_{\text{lat}}),\quad 
           \eta_{\text{lon}}^{(k)} \sim \mathrm{Beta}(a_{\text{lon}}, b_{\text{lon}})$
    \EndFor
    
    \State \textcolor{gray}{\textbf{// Step 5: Guided denoising.}}
    \For{$k = 1$ to $G_{\text{grpo}}$}
        \State $x^{(k)}_{S} \gets z$, $z \sim \mathcal{N}(\mathbf{0}, \mathbf{I})$
        \For{$i = 1$ to $S$}
            \State $s \gets s_i$
            \State {\small$
                x^{(k)}_{s-1} 
                \gets 
                D_{\theta}(
                    x^{(k)}_{s},
                    s,
                    F_{\text{scene}},
                    F_{\text{navi}},
                    \eta_{\text{lat}}^{(k)},
                    \eta_{\text{lon}}^{(k)},
                    x^{\text{ref}}
                )
            $} \Comment{classifier-guided denoising following \cref{eq:guided}}
        \EndFor
        \State $x^{(k)} \gets x^{(k)}_{0}$
    \EndFor
    
    \State \Return $\{x^{(k)}\}_{k=1}^{G_{\text{grpo}}}$  \Comment{Multi-modal trajectory samples}
    \end{algorithmic}
\end{algorithm}

\boldparagraph{Q3.}
\textit{What are potential applications and future directions with the PlannerRFT framework and the nuMax simulator?}

\textbf{Model}: Build upon a simple diffusion planner, PlannerRFT can enhance the multi-modality and adaptive sampling in RL. 
We freeze the encoder and fine-tune only the trajectory DiT, which suggests a potential ability to act as a unified decoder for different input modalities, such as sensor-based E2E planners or language-conditioned VLM/VLA planners.

\textbf{Simulator}: To support our training pipeline, we develop nuMax, a fast online RL simulator designed to facilitate academic research on the nuPlan benchmark. Additional implementation details, limitations, and future development plans are provided in \cref{sec:numax}.

\section{Implementation Details of PlannerRFT}

\boldparagraph{Training Details.} \cref{fig:ddp} show the training pipeline for the PlannerRFT framework. For the commonly used online RL, the training pipeline involves two steps: (1) RL sampling and (2) policy update. 

\textbf{For RL sampling}, PlannerRFT adopt the policy-guide denoising to generate multi-modal and scenario-adaptive trajectocy group.
\cref{alg:rl_sampling} outlines the details of the policy-guide denoising process. We adopt a 5 steps DDIM sampling during training and inference for computational efficiency and exploration stochasticity. 
\begin{equation}
x_{s-1} = \sqrt{\alpha_{s-1}}\, \hat{x}_{s}^{0} 
+ \sqrt{1-\alpha_{s-1}-\sigma_s^2}\, \epsilon_\theta(x_s,s) 
+ \sigma_s z
\end{equation}
\begin{equation}
\epsilon_\theta(x_s,s) = 
\frac{x_{s}-\sqrt{\alpha_{s}\, \hat{x}_{s}^{0}}}{\sqrt{1-\alpha_{s}}}
\end{equation}
\begin{equation}
\sigma_s = 
\eta \cdot \sqrt{\frac{1-\alpha_{s-1}}{1-\alpha_s}},
\end{equation}
where $s$ is the denoising timestep, $\hat{x}_{s}^{0}$ is the model-predicted clean trajectory at timestep $s$, $\sigma_s$ controls the stochasticity of DDIM sampling, and
$z\sim\mathcal{N}(0,I)$ denotes standard Gaussian noise. Specifically, we set $\eta=1$ during RL training to encourage stochastic exploration,
and $\eta=0$ during evaluation for deterministic sampling.

\textbf{For policy update}, PlannerRFT consists of two learnable modules: the exploration policy and the fine-tuned DiT, which have different optimization goals and are trained with different optimization losses.

We use the PPO loss to update the exploration policy, aiming to maximize the long-term cumulative reward in closed-loop planning.
\begin{equation}
\begin{aligned}
    \mathcal{L}_{\text{PPO}}(\phi)
    =
    \mathbb{E}_t\Big[
    & \mathcal{L}_\text{clip}(\phi )
    -
    c_v\, (V_\phi(s_t) - V_t^{\text{target}})^2 \\
    & + c_e\, \mathcal{H}\!\left(\pi_\phi(\cdot|o_t)\right)
    \Big]
\end{aligned}
\end{equation}
\begin{equation}
\begin{aligned}
    \mathcal{L}_\text{clip}(\phi )
    &= \min\Big(
        r_t(\phi) A_t, \\
    &\qquad\quad
        \mathrm{clip}\big(r_t(\phi), 1-\epsilon, 1+\epsilon\big) A_t
    \Big)
\end{aligned}
\end{equation}
\begin{equation}
    r_t(\phi)=
    \frac{\pi_\phi(\eta_t \mid o_t)}
    {\pi_{\phi{\text{old}}}(\eta_t \mid o_t)},
\end{equation}
where $\mathcal{L}_{\text{clip}}$ is the clipped policy objective, $r_t(\phi)$ is the importance sampling ratio, $A_t$ is the advantage estimate, $V_\phi(s_t)$ is the value prediction, $\mathcal{H}(\pi_\phi)$ denotes the entropy bonus, and $c_v, c_e$ are the value and entropy coefficients. These hyperparameters are summarized in \cref{tab:hyperparameter}.

\begin{table}[t]
\centering
\caption{
    \textbf{Hyperparameters for PlannerRFT.}
}
    \vspace{-6pt}
\scalebox{0.9}{
{\setlength{\tabcolsep}{8pt}
    \begin{tabular}{@{}c|ll@{}}
    \toprule
    \multicolumn{2}{c}{\textbf{Hyperparameter}}     & \textbf{Value}         \\ \midrule
    \multirow{2}{*}[0.5ex]{\makecell[c]{Guidance }}  & Max. Lateral Offset $\lambda_{\text{lat.}}$       & $2.5 ~ (m)$         \\  
    & Max. Longitudinal Offset $\lambda_{\text{lon.}}$       & $25 ~ (\%)$         \\ \midrule
    \multirow{14}{*}[0.5ex]{\makecell[c]{PPO }} & Samples  & 40M     \\
    & Initial Learning Rate       & $2.5\times10^{-4}$         \\
    & Learning Rate Schedule      & Cosine decay         \\
    & Number of Envs.             & 128         \\
    & Env. Steps per Iteration    & 32            \\
    & Batch Size                  & 4096            \\
    & Mini-batch Size             & 4096            \\
    & Steps per Epoch             & 1            \\
    & Epochs                      & 4            \\
    & Value Coefficient $c_{v}$   & 0.5            \\
    & Entropy Coefficient $c_{e}$ & 0.01            \\
    & Discount Factor             & 0.99            \\
    & GAE $\lambda$               & 0.95            \\
    & Clip Range $\epsilon$       & 0.2            \\
    & Max Gradient Norm           & 0.5           \\ \midrule
    \multirow{8}{*}[0.5ex]{\makecell[c]{GRPO }} & Initial Learning rate  & $2.5\times10^{-4}$     \\
    & Learning Rate Schedule      & Cosine decay         \\
    & Group Size $G_{grpo}$       & 8           \\
    & Mini-batch Size             & 4096        \\
    & Steps per Epoch             & 6           \\
    & Epochs                      & 1            \\
    & Denoising Discount Factor $\gamma$ & 0.8          \\
    & BC loss weight $c_{b}$     & 0.4 \\
    \bottomrule
    \end{tabular}
}}
\label{tab:hyperparameter}
    \vspace{-3pt}
\end{table}

\begin{table}[t]
\centering
\caption{
    \textbf{Comparison of different inference types.}
    ``$\text{Diffusion Planner}_{\texttt{DPM}}$'' is the official 10-step DPM-solver version of Diffusion Planner~\cite{DiffusionPlanner}.
    ``w/ guid.'' denotes inference with guided denoising, where the guidance scale is set to the mean of the Beta distribution.
    ``w/o guid.'' denotes inference without guidance.
}
    \vspace{-6pt}
\scalebox{0.8}{
{\setlength{\tabcolsep}{5pt}
    \begin{tabular}{@{}l|cc|cc@{}}
    \toprule
    Model & Steps & Latency ($ms$) & Val14-NR & Val14-R \\ \midrule
    $\text{Diffusion Planner}_{\texttt{DPM}}$ & 10 & 86.43 & 89.87 & 82.80 \\
    PlannerRFT w/ guid.  & 10 & 75.48 & 89.83 & 83.93 \\
    PlannerRFT w/o guid. & 5  & \textbf{34.27} & \textbf{89.96} & \textbf{84.46} \\
    \bottomrule
    \end{tabular}
}}
\label{tab:inference}
    \vspace{-15pt}
\end{table}

We use the GRPO loss to update the fine-tuned DiT, aiming to maximize the reward within the prediction horizon at the current timestep. Following DPPO~\cite{DPPO}, each conditional step in the diffusion chain is a Gaussian policy:
\begin{equation}
    \pi_{\theta}(x_{s-1} \mid x_s)
    = \mathcal{N}\!\Big(x_{s-1};\; \mu_{\theta}(x_{s}, s),\; \sigma_s^{2} I
    \Big)
\end{equation}
\begin{equation}
    \begin{aligned}
    \mu_{\theta}(x_{s}, s)= &\sqrt{\alpha_{s-1}}\, \hat{x}^{0}_{s} \\
            &+ \sqrt{1-\alpha_{s-1}-\sigma_s^2}\, \epsilon_\theta(x_s,s),
    \end{aligned}
\end{equation}
where $\mu_{\theta}(x_s, s)$ is the deterministic update term in DDIM, and
$\sigma_s^2 I$ controls the sampling stochasticity. Therefore, the optimization objective is to adjust the denoising process such that the conditional policy $\pi_{\theta}(x_{s-1}\mid x_s)$ is shifted toward trajectories with higher expected rewards:
\begin{equation}
\begin{aligned}
\mathcal{L}_{G}
&=-
\frac{1}{G_{grpo}} \sum_{k=1}^{G_{grpo}} \frac{1}{S} \sum_{s=1}^{S}
\gamma^{s-1} \log \pi_{\theta}\!\left(x^{(k)}_{s-1} \mid x^{(k)}_{s}\right)\, \hat{A}_k
\\[6pt]
&\quad
-\;c_{b}\,
\frac{1}{G_{grpo}} \sum_{k=1}^{G_{grpo}} \frac{1}{S} \sum_{t=1}^{S}
\log \pi_{\theta}\!\left(\tilde{x}^{(k)}_{s-1} \mid \tilde{x}^{(k)}_{s}\right),
\end{aligned}
\end{equation}
where $\gamma$ is the denoising discount factor. Following ReCogDrive~\cite{ReCogDrive}, we incorporate a behavior cloning loss to prevent policy collapse during exploration, and $c_b$ denotes the weight of the behavior cloning term.

\boldparagraph{Inference Details.} 
For inference, we adopt the same 5-step DDIM sampler as in the training phase, without guided denoising or reliance on the reference planner.
\Cref{tab:inference} compares different inference settings in terms of denoising steps, latency, and closed-loop performance on the Val14 benchmark.
PlannerRFT with guided denoising requires twice as many denoising steps because each step depends on the reference planner’s trajectory. This additional dependency also results in nearly double the inference latency compared to the unguided version.
Performance-wise, the guided version is slightly worse than PlannerRFT without guidance.
Exploration guidance provides a directional prior that enables sampling around the intended exploration direction, yielding both positive and negative trajectory examples that help refine the learned distribution. Under a limited training budget, however, the model may not fully capture an accurate distribution mean, particularly in fine-grained maneuvering scenarios, leading to slight performance degradation. These effects collectively highlight the sampling efficiency of PlannerRFT.

\begin{figure*}[t]
    \centering
    \includegraphics[width=\linewidth]{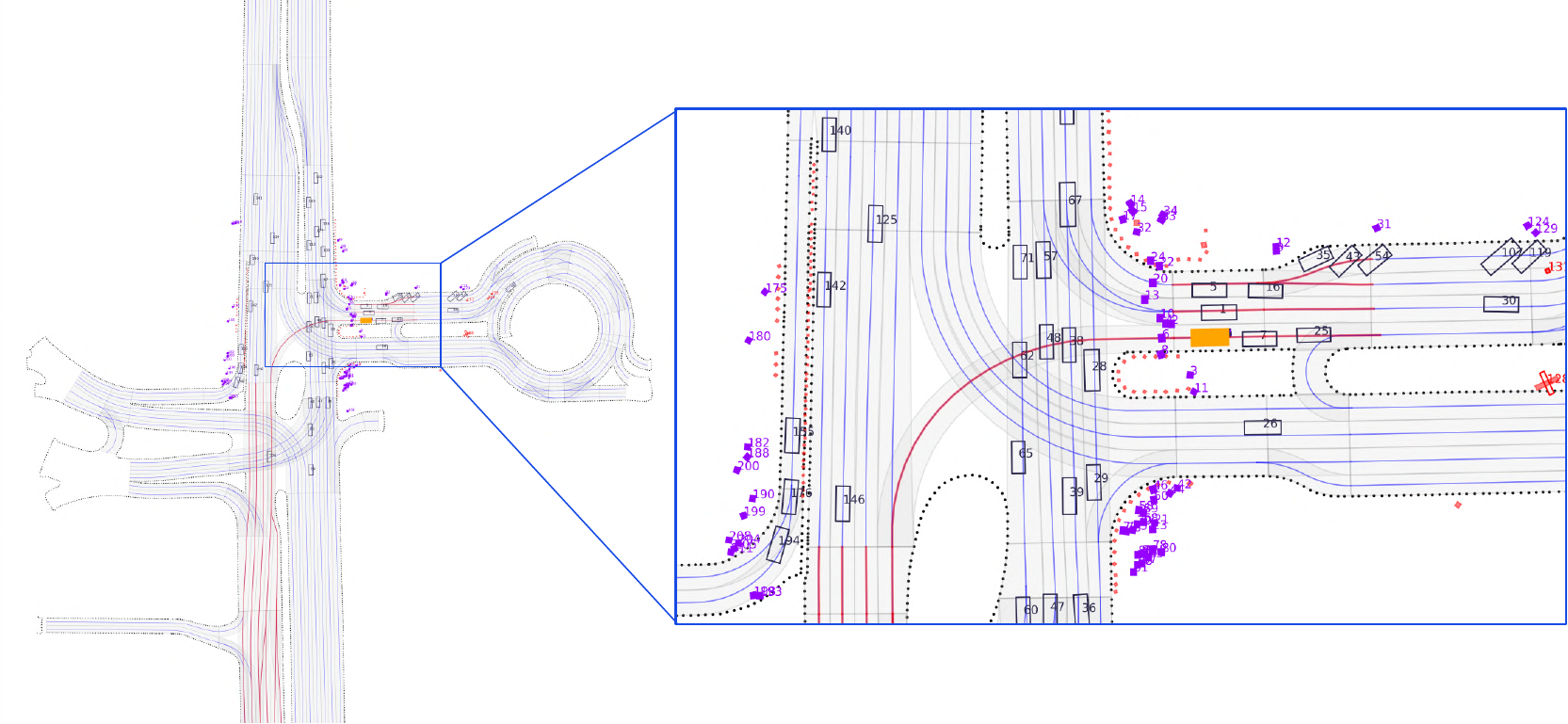}
    \caption{
    \textbf{Visualization of a Cached Scenario.} 
    We cache scene elements within a 200~$m$ radius of the \textcolor{orange}{ego vehicle}, including lanes, dynamic agents, static obstacles, and navigation routes. 
    \textcolor{gray}{Lane polygons} are drawn in gray with blue \textcolor{blue}{centerlines}, while \textcolor{red}{navigation routes} are highlighted in red. 
    \textcolor{black}{Surrounding vehicles} are drawn as black rectangles, 
    \textcolor[HTML]{9500ff}{pedestrians} and \textcolor[HTML]{9500ff}{cyclists} in purple, and \textcolor{red}{static objects} in red. 
    }
    \label{fig:scenario_cache}
    \vspace{-6pt}
\end{figure*}

\section{Implementation Details of nuMax} \label{sec:numax}

We develop nuMax, a JAX-based, GPU-parallel simulator built upon Waymax~\cite{Waymax}, to support large-scale closed-loop training on nuPlan.
nuMax achieves significantly higher simulation speed compared to the official nuPlan simulator by re-designing the data representation, scene update pipeline, and agent dynamics entirely around JAX’s functional. Below, we summarize the implementation details.

\boldparagraph{Scenario Cache.}
Efficient high-throughput data loading is crucial for large-scale training in closed-loop simulation. 
In the official nuPlan dataset, scenario recordings are stored in an SQLite\footnote{\url{https://sqlite.org}} database and HD maps reside in a GeoPandas\footnote{\url{https://geopandas.org}} dataframe, requiring the simulator to query both sources at every simulation step to retrieve lane geometry, dynamic agents, and scene context. 
This stepwise database access limits the simulation throughput. 
Consequently, nuMax pre-caches the training scenario based on ScenarioMax\footnote{\url{https://github.com/valeoai/ScenarioMax}}, a high-performance toolkit for autonomous vehicle scenario-based testing and dataset conversion.
Specifically, for temporal context, we extract a fixed window consisting of 20 past frames, the current frame, and 150 future frames at a sampling interval of 0.1 s.
For spatial context, we crop all scene elements within a 200 m radius centered on the ego vehicle, including lanes, dynamic agents, static obstacles, and navigation routes.
An example visualization of the scenario cache is shown in \cref{fig:scenario_cache}.
All processed data are serialized into TFRecord\footnote{\url{https://www.tensorflow.org/tutorials/load_data/tfrecord}} files, enabling fast sequential I/O, efficient GPU loading, and full compatibility with JAX’s parallelized execution model, thereby eliminating runtime database queries and supporting high-throughput simulation in nuMax.

\boldparagraph{Tracker and Scorer.}
Reliable vehicle motion tracking is essential for robust closed-loop simulation. 
Waymax adopts a vehicle controller built on Perfect Control, but we found it occasionally understeers in sharp-turn scenarios.
In nuMax, we replace the perfect-control controller with the two-stage motion controller from the official nuPlan-devkit\footnote{\url{https://github.com/motional/nuplan-devkit}}
, which consists of an LQR tracker and a kinematic bicycle model.
Our implementation follows the controller used in PDM-Closed\footnote{\url{https://github.com/autonomousvision/tuplan_garage}}
, which extends the controller to support batched trajectory inputs, enabling nuMax to track an entire group of candidate trajectories simultaneously during rollouts.
We reimplement the two-stage controller in JAX and integrate it into nuMax’s GPU-parallel simulation pipeline, where XLA\footnote{\url{https://openxla.org/xla}} compilation further accelerates tracking and improves overall computational efficiency.

Comprehensive and principled reward evaluation is essential for effective reinforcement learning policy optimization.
Building upon the metrics provided by Waymax, we further incorporate the official nuPlan scoring framework. In particular, our scorer includes both terminal penalties and soft penalties, enabling a more complete assessment of driving quality and safety during closed-loop rollouts.

For terminal penalties, once the ego violates any terminal condition, the simulation is immediately terminated and the reward is set to zero. The terminal penalties include:
\begin{itemize}
\item \textbf{Collision (\texttt{Col})}: if the ego vehicle collides with surrounding vehicles, pedestrians, cyclists, or static objects.
\item \textbf{Off road (\texttt{DAC})}: if the ego vehicle drove off the drivable area.
\end{itemize}

For soft penalties, the ego aims to minimize these penalties while avoiding any terminal violations. The soft penalties include:
\begin{itemize}
\item \textbf{Wrong direction (\texttt{WD})}: if the ego vehicle drives against the designated lane direction.
\item \textbf{Time to collision (\texttt{TTC})}: if the ego vehicle violates the time-to-collision (TTC) safety threshold.
\item \textbf{Comfort (\texttt{C})}: if the ego vehicle exhibits excessive longitudinal/lateral acceleration, jerk, or steering rate.
\item \textbf{Ego Progress (\texttt{EP})}: measured as the normalized ratio between the ego’s accumulated route progress and that of the expert.
\item \textbf{Speeding (\texttt{Speed})}: if the ego vehicle exceeds the speed limit of the current lane or route segment.
\end{itemize}

Each component score lies within $[0,1]$, and the final reward is obtained by weighted aggregation of all metrics:
\begin{equation}
\begin{split}
    r_t &= (\texttt{Col} \cdot \texttt{DAC} \cdot \texttt{WD}) \\
        &\times 
        \left(
            \frac{
                w_{1}\texttt{TTC} +
                w_{2}\texttt{EP} +
                w_{3}\texttt{C} +
                w_{4}\texttt{Speed}
            }{
                \sum_i w_i
            }
        \right)
        \in [0, 1],
\end{split}
\end{equation}
where we adopt the official nuPlan weights: $w_{1} = w_{2} = 5.0$, $w_{3}=2.0$, and $w_{4}=4.0$.
For the open-loop scorer used for GRPO trajectory evaluation, we adopt the same metric terms, but replace the terminal-penalty computation with the survival formulation defined in \cref{eq:survival}.

\begin{figure}[t]
    \centering
    \includegraphics[width=\linewidth]{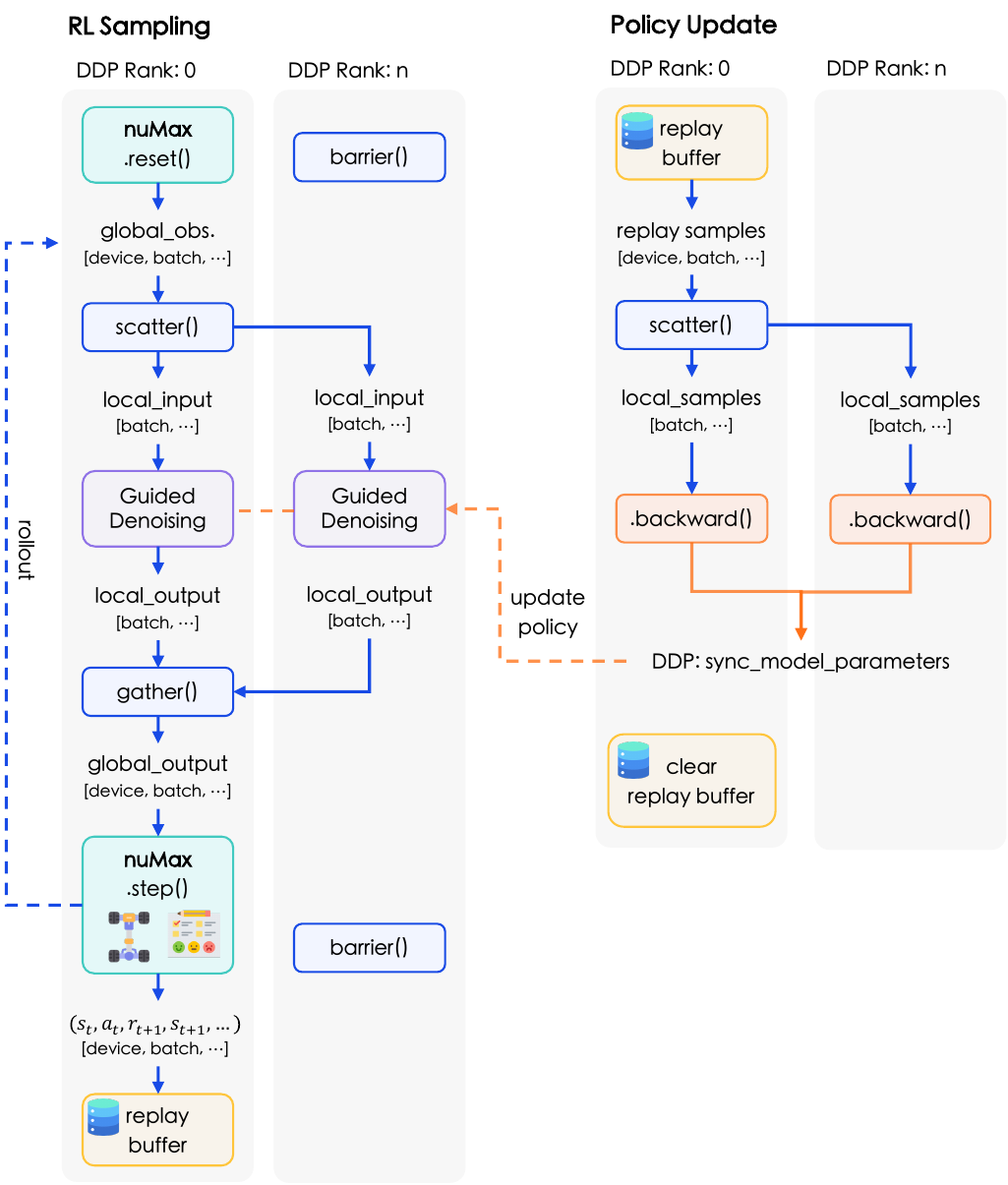}
    \caption{
        \textbf{Distributed RL Training Pipeline.}
        Policy inference and learning run in PyTorch DDP, while environment simulation is executed in JAX on rank-0 to avoid XLA conflicts. Observations and replay samples are scattered to all ranks for guided denoising and gradient updates, and planned trajectories are gathered back to rank-0 for simulation, with synchronization at every step.
    }
    \label{fig:ddp}
\end{figure}

\boldparagraph{Distributed RL Training Pipeline.}
For the reinforcement learning pipeline, we refer to V-Max~\cite{vmax},
a JAX-based high-performance framework built on the Brax\footnote{\url{https://github.com/google/brax}} engine, which integrates simulation pipelines, observation wrappers, and evaluation metrics.
However, most diffusion-based planners widely adopted in the community are implemented in PyTorch, 
and re-implementing the entire model stack in JAX would be both costly and incompatible with existing pretrained models. 
To preserve compatibility and facilitate broader community adoption, we therefore retain the policy model in PyTorch.
Based on these considerations, we design a hybrid distributed RL training pipeline that couples JAX-based simulation with PyTorch Distributed Data Parallel (DDP) for policy inference and learning.

\cref{fig:ddp} illustrates our hybrid distributed RL training pipeline. All JAX-based simulation is executed on rank-0 to avoid XLA device conflicts and duplicate backend initialization that would arise from launching independent JAX runtimes on each PyTorch DDP rank. We distribute the observations and replay samples to all DDP ranks, and aggregate the planned trajectories back to rank~0 for simulation. In addition, we ensure that all ranks are synchronized before each simulation step.

\boldparagraph{Limitations of nuMax.}
nuMax currently inherits two key limitations.
First, due to XLA’s static-shape constraints, supporting other model input representations would require additional post-processing or re-caching, highlighting the need for a general scenario cache interface.
Second, surrounding-vehicle simulation is log-replay rather than IDM, as the latter slows down training; improving the efficiency of the IDM traffic simulation remains an important direction for future optimization.

\begin{table*}[t]
\centering
\caption{
    \textbf{Closed-loop Planning Results on nuPlan Val14, Test14-hard, and Test14 benchmarks.}
    ``$\text{Diffusion Planner}_{\texttt{DPM}}$'' employs the official 10-step DPM-solver.  
    ``$\text{Diffusion Planner}_{\texttt{DDIM}}$'' employs a 5-step DDIM sampler, identical to that used in PlannerRFT.
}
    \vspace{-6pt}
\scalebox{0.9}{
{\setlength{\tabcolsep}{5pt}
    \begin{tabular}{@{}llccccccccccc@{}}
    \toprule
    \multirow{2}{*}[-0.15ex]{\makecell[l]{\textbf{Type}}}     & \multirow{2}{*}[-0.15ex]{\makecell[l]{\textbf{Planner}}}       & \multicolumn{2}{c}{\textbf{Val14}} & \multicolumn{2}{c}{\textbf{Test14-hard}} & \multicolumn{2}{c}{\textbf{Test14-random}} \\ 
    \cmidrule(lr){3-4} \cmidrule(lr){5-6} \cmidrule(lr){7-8}
    &  & \textbf{NR} & \textbf{R} & \textbf{NR} & \textbf{R} & \textbf{NR} & \textbf{R} \\ \midrule
    \textcolor{gray}{Expert} & \textcolor{gray}{Log-replay} & \textcolor{gray}{93.53} & \textcolor{gray}{80.32} & \textcolor{gray}{85.96} & \textcolor{gray}{68.80} & \textcolor{gray}{94.03} & \textcolor{gray}{75.86} \\ \midrule
    \multirow{2}{*}[-0.15ex]{\makecell[l]{Rule}}
    & IDM         & 75.60  & 77.33  & 56.15  & 62.26 & 70.39  & 74.42 \\ 
    & PDM-Closed  & 92.84  & 92.12  & 65.08  & 75.19 & 90.05  &  91.63 \\ 
    \midrule
    \multirow{8}{*}[0.0ex]{\makecell[l]{Learning }}
    & PDM-Open            & 53.53  & 54.24  & 33.51  & 35.83  & 52.81  & 57.23  \\  
    & GameFormer          & 13.32  & 8.69   & 7.08   & 6.69   & 11.36  & 9.31 \\ 
    & PlanTF              & 84.27  & 76.95  & 69.70  & 61.61  & 85.62  & 79.58\\ 
    & PLUTO               & 88.89  & 78.11  & 70.03  & 59.74  & \underline{89.90}  & 78.62\\ 
    & $\text{Diffusion Planner}_{\texttt{DPM}}$   & 89.87  & 82.80  & 75.99  & 69.22 & 89.19  & \underline{82.93}\\
    & $\text{Diffusion Planner}_{\texttt{DDIM}}$   & \colorbox{light}{89.81}  & \colorbox{light}{82.94}  & \colorbox{light}{76.01}  & \colorbox{light}{68.18} & \colorbox{light}{89.14} & \colorbox{light}{82.63}\\
    & Flow Planner        & \colorbox{mine}{\textbf{90.43}}  & \underline{83.31}  & \underline{76.47} & \underline{70.42} & 89.88 & \underline{82.93} \\
    & \textbf{PlannerRFT(Ours)}  & \underline{89.96{\scriptsize \textcolor{blue}{(+0.15)}}}  & \colorbox{mine}{\textbf{84.46{\scriptsize \textcolor{blue}{(+1.52)}}}}  & \colorbox{mine}{\textbf{77.16{\scriptsize \textcolor{blue}{(+1.15)}}}} & \colorbox{mine}{\textbf{72.21{\scriptsize \textcolor{blue}{(+4.03)}}}}  & \colorbox{mine}{\textbf{90.76{\scriptsize \textcolor{blue}{(+1.62)}}}} & \colorbox{mine}{\textbf{85.80{\scriptsize \textcolor{blue}{(+3.17)}}}} \\
    \bottomrule
    \end{tabular}
}}
\label{tab:testr}
\end{table*}

\section{Additional Ablation Studies}

\begin{table*}[htbp!]
\caption{
    \textbf{Ablation on Guidance Type.}
    Results are reported on the Test14-random reactive benchmark.
    }
\vspace{-6pt}
\centering
\scalebox{0.9}{
    \begin{tabular}{l|cc|cccccc|c}
        \toprule
        & \multicolumn{2}{c|}{\textbf{Guidance Choices}} 
        & \multicolumn{7}{c}{\textbf{Closed-loop metrics $\uparrow$}} \\ \midrule

        Training Type & lateral & longitudinal & Collisions & TTC & Drivable & Comfort & Progress & Speed & \textbf{R-score} \\
        \midrule

        $\text{IL Pretrain}_\texttt{DDIM}$ & \xmark & \xmark & 86.58 & 79.05 & 94.48 & 86.03 & 76.99 & 97.20 & 68.18 \\ \midrule

        \multirow{3}{*}[0.0ex]{\makecell[l]{PlannerRFT }} 
        & \xmark & \cmark & \underline{87.50} & \underline{81.62} & 92.65 & \underline{86.76} & \textbf{77.54} & \underline{97.99} & 69.59 \\
        
        & \cmark & \xmark  & 87.31 & 80.88 & \underline{94.85} & \textbf{87.50} & 76.38 & 97.32 & \underline{70.18} \\
        
        & \cmark & \cmark & \textbf{88.97} & \textbf{84.93} & \textbf{95.59} & 85.66 & \underline{77.17} & \textbf{98.03} & \textbf{72.21} \\
        
        \bottomrule
    \end{tabular}
}
    \label{tab:abl_guidance}
\end{table*}

\boldparagraph{Additional planning results in Test14-Random.}
Table~\ref{tab:testr} reports the closed-loop performance on the Test14-random benchmark, which contains 261 randomly selected scenarios from the nuPlan Planning Challenge. As shown in \cref{tab:testr}, the 5-step DDIM sampler and the ODE-based DPM-solver yield nearly identical results, ensuring a fair comparison. PlannerRFT achieves the best performance in both non-reactive (NR) and reactive (R) settings in Test14-random, improving over the pretrained planner by +1.62 (NR) and +3.17 (R).

\boldparagraph{Ablation on Guidance Choices.}
We evaluate the effectiveness of lateral and longitudinal guidance by enabling them individually in the policy-guided denoising process. As shown in \cref{tab:abl_guidance}, lateral guidance improves performance in Drivable and Comfort, as it enhances sharp-turn performance and produces smoother lateral maneuvers. In contrast, longitudinal guidance performs better in terms of Collisions, TTC, Progress, and Speed, as these behaviors can be controlled through acceleration and deceleration. Combining both forms of guidance results in the best performance, highlighting the complementary effect of lateral and longitudinal exploration in optimizing closed-loop planning.

\boldparagraph{Ablation on Group Size.}
We evaluate the impact of group size on performance by testing three different values of $G_{\text{grpo}}$ in the Test14-hard benchmark. As shown in \cref{tab:abl_group_size}, using a group size of 4 results in suboptimal performance compared to larger group sizes. When the group size is increased to 8 or 12, performance improves, with scores stabilizing around 72.21 (R-score) and 77.16 (NR-score). We choose a group size of 8 to strike an optimal balance between performance and computational efficiency.

\begin{table}[t]
\caption{
    \textbf{Ablation on Group number.}
    Results are reported on the Test14-random benchmark.
    }
\vspace{-6pt}
\centering
\scalebox{0.9}{
    {\setlength{\tabcolsep}{5.5pt}
    \begin{tabular}{@{}c|lcccc|l@{}}
    \toprule
    \multirow{2}{*}[0.0ex]{\makecell[l]{$G_{grpo}$}} & \multicolumn{6}{c}{\textbf{Closed-loop metrics $\uparrow$}} \\ \cmidrule(lr){2-7}
     & \textbf{R-score} & Coll. & Driv. & C. & Prog. & \textbf{NR-score}   \\ \midrule
    4  & 71.24 & \underline{86.40} & 94.85 & \underline{84.93} & \textbf{78.99} & 76.31 \\
    8  & \underline{72.21} & \textbf{88.97} & \textbf{95.59} & \textbf{85.66} & 77.17 & \textbf{77.16} \\
    12 & \textbf{72.29} & \textbf{88.97} & \underline{95.22} & \textbf{85.66} & \underline{77.62} & \underline{77.04} \\
    \bottomrule
    \end{tabular}
}}
\label{tab:abl_group_size}
\vspace{-6pt}
\end{table}

\section{Additional Qualitative Results}

\begin{figure*}[t]
    \centering
    \includegraphics[width=\linewidth]{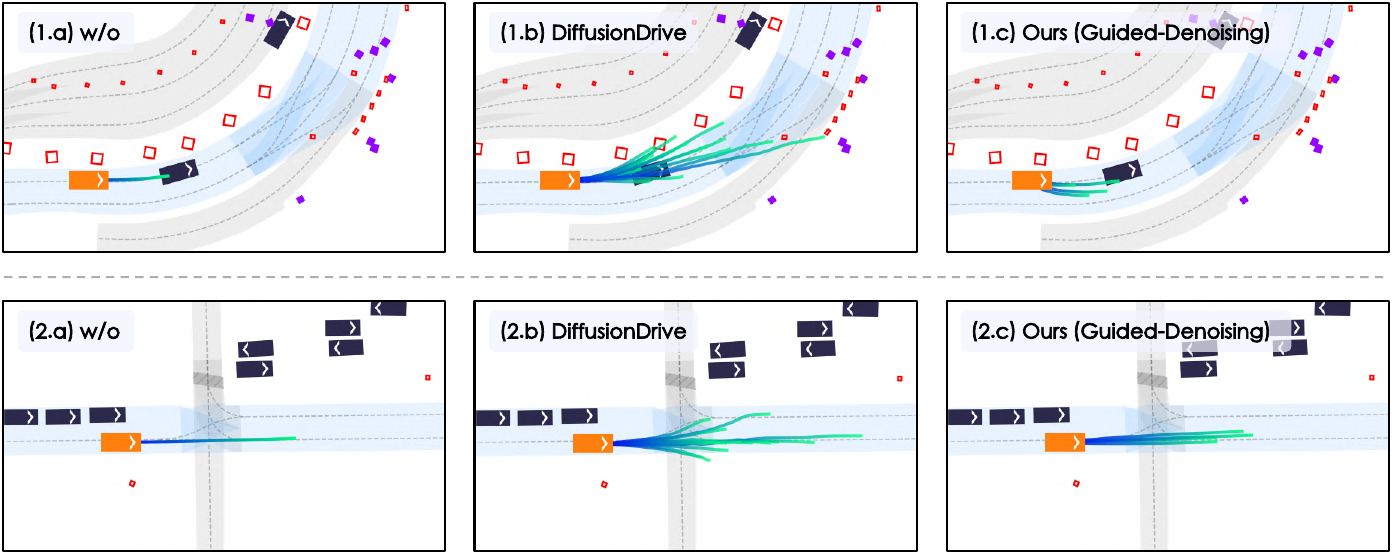}
    \caption{
        \textbf{Visualization of Diffusion Planner (a, w/o guided denoising), DiffusionDrive (b) and PlannerRFT (c, w/ guided denoising).} 
        For Diffusion Planner and PlannerRFT, we resample 4 trajectories, for DiffusionDrive we use 20 anchor noises. We visualize the planned trajectory over a 4 $second$ horizon.
        Note that DiffusionDrive is evaluated on the NAVSIM \texttt{navtest} split with camera and LiDAR inputs; for visualization, we render all planners on the same scenario shared between NAVSIM and nuPlan. PlannerRFT demonstrates multi-modal and scenario-adaptive trajectory generation through policy-guided denoising.
    }
    \label{fig:sup_sampling}
\end{figure*}

\boldparagraph{Additional Visualization of Policy-Guided Denoising.}
We show qualitative comparisons of planned trajectories from Diffusion Planner, DiffusionDrive, and PlannerRFT with policy-guided denoising, as shown in \cref{fig:sup_sampling}. With policy-guided denoising, PlannerRFT generates a group of multi-modality and scenario-adaptive trajectories for sampling efficient RL training.

\begin{figure*}[t]
    \centering
    \includegraphics[width=\linewidth]{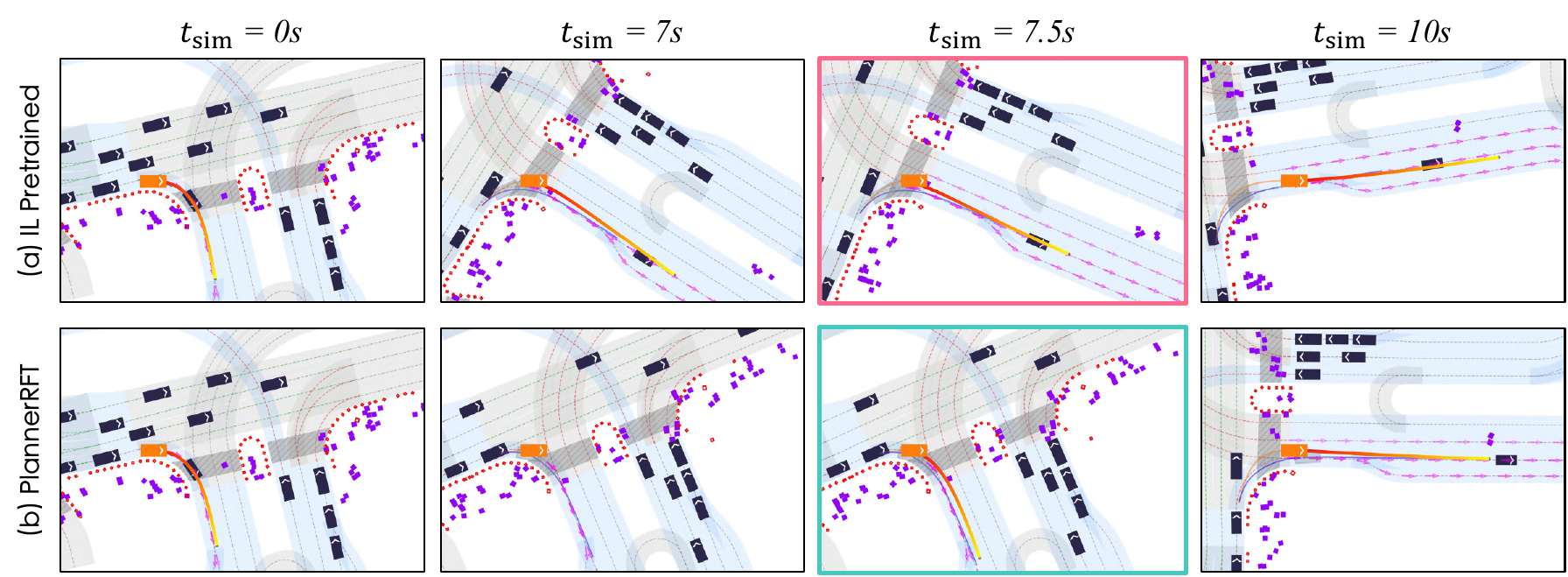}
    \caption{
        \textbf{Intersection Pedestrian Avoidance.} 
        The ego vehicle intends to make a right turn at an intersection while pedestrians are crossing. 
        \textbf{(a)} The IL Pretrained planner collision with a pedestrian at $t_{\text{sim}}=7.5~s$. 
        \textbf{(b)} PlannerRFT waits for all pedestrians to finish crossing and then proceeds with the right turn.
        In each frame shot, the \textcolor{orange}{simulation position} and \textcolor{orange}{planning trajectory} are marked as orange, the \textcolor{gray}{ground-truth position} and \textcolor{blue}{ground-truth trajectory} recorded in the driving log are marked as gray and blue, respectively. Surrounding vehicles are marked as black rectangles with white arrows indicating heading. The \textcolor[HTML]{9500ff}{pedestrians} are marked as purple, and the \textcolor{red}{static objects} are marked as red. The \textcolor{gray}{lane polygons} are marked gray and the \textcolor[HTML]{1E90FF}{navigation routes} are marked as light blue with \textcolor{magenta}{centerline arrows}.
    }
    \label{fig:demo_safety1}
\end{figure*}

\begin{figure*}[t]
    \centering
    \includegraphics[width=\linewidth]{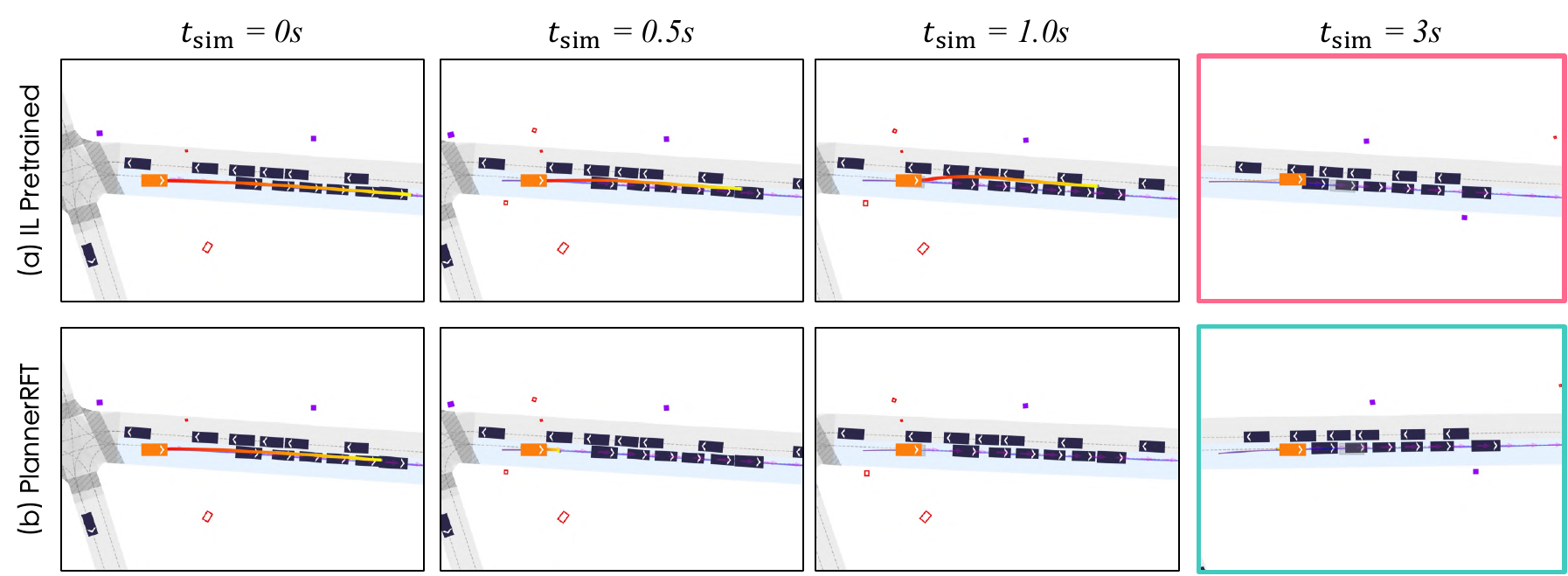}
    \caption{
        \textbf{Emergency Brake in \textcolor{blue}{Reactive Traffic}.} 
        A safety-critical scenario in which surrounding vehicles governed by the IDM policy enter a deadlock at the initial timestep ($t_{\text{sim}}=0~s$), blocking all traffic, while the ego vehicle approaches at high speed as recorded in the log. 
        \textbf{(a)} The IL-pretrained planner fails to brake in time and collides with the preceding vehicle.
        \textbf{(b)} PlannerRFT detects the stationary lead vehicle and applies braking at $t_{\text{sim}}=1~s$, successfully avoiding the collision.
    }
    \label{fig:demo_safety2}
\end{figure*}

\boldparagraph{Qualitative Results on Safety-Critical Scenarios.}
We present closed-loop planning results of PlannerRFT on safety-critical scenarios, as shown in \cref{fig:demo_safety1} and \cref{fig:demo_safety2}. These examples demonstrate the planner’s enhanced safety awareness, improved maneuver robustness, and better handling of dynamic interactions.

\begin{figure*}[t]
    \centering
    \includegraphics[width=\linewidth]{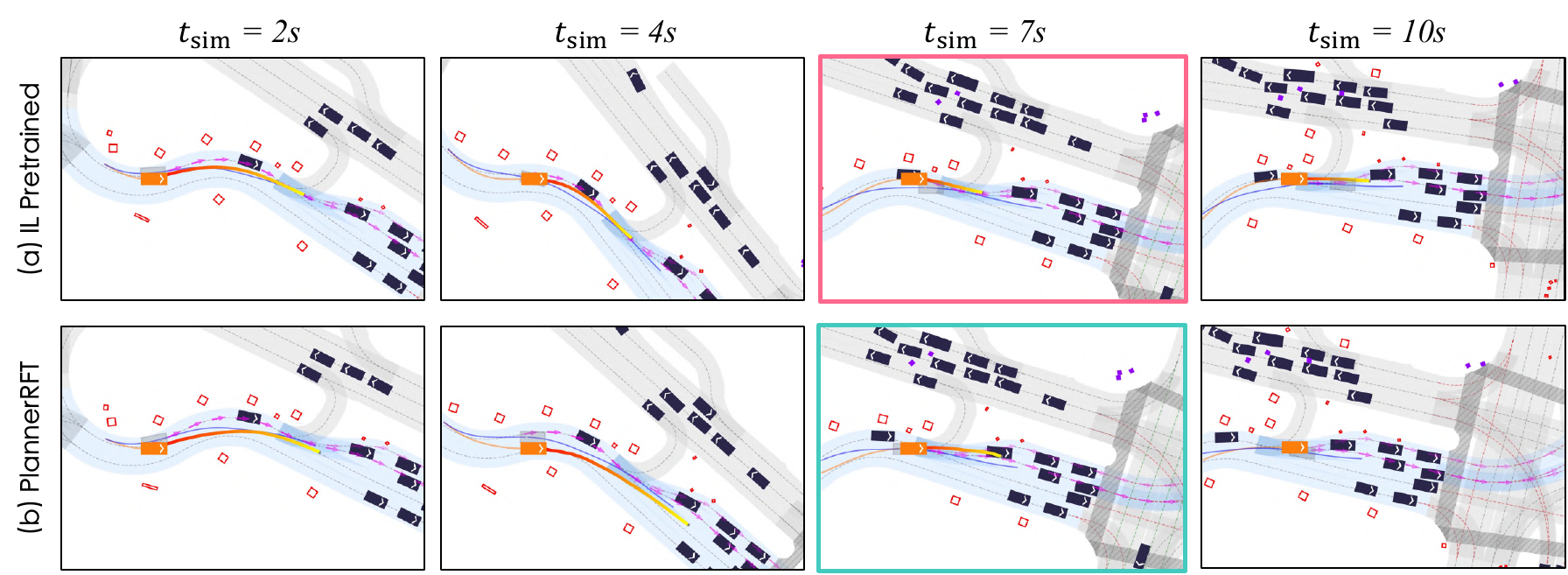}
    \caption{
        \textbf{S-Curve Lane Change.} 
        The ego vehicle starts in the left lane of an S-curve, with a stationary vehicle ahead in the same lane.
        \textbf{(a)} The IL-pretrained planner keeps the lane and collides with the stationary vehicle at $t_{\text{sim}}=7~s$.
        \textbf{(b)} PlannerRFT performs a lane change to the right at $t_{\text{sim}}=4~s$, bypassing the stationary vehicle.
    }
    \label{fig:demo_obstacle1}
\end{figure*}

\begin{figure*}[t]
    \centering
    \includegraphics[width=\linewidth]{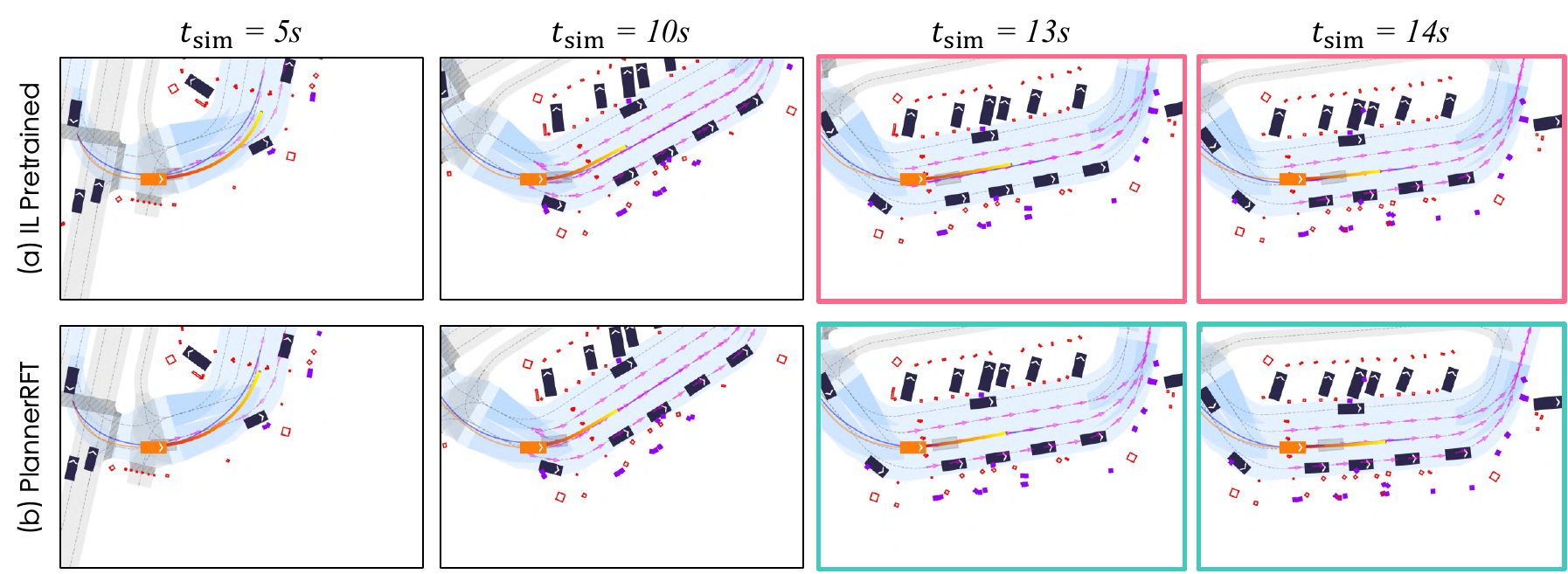}
    \caption{
        \textbf{Traffic-Cone Narrowing.} 
        The ego vehicle is driving on a curved road with traffic cones. 
        \textbf{(a)} The IL Pretrained planner based ego vehicle fails to avoid the traffic cone in time, colliding with it at $t_{\text{sim=13}}s$. 
        \textbf{(b)} PlannerRFT enables the ego vehicle to finely adjust the trajectory, successfully steering the ego vehicle between the two cones.
    }
    \label{fig:demo_obstacle2}
\end{figure*}

\boldparagraph{Qualitative Results on Obstacle Avoidance Scenarios.}
We present the closed-loop planning results for PlannerRFT on obstacle avoidance  scenarios, as shown in \cref{fig:demo_obstacle1} and \cref{fig:demo_obstacle2}.
These examples demonstrate the planner’s ability to laterally avoid obstacles and to execute precise maneuvers in narrow spaces.

\begin{figure*}[t]
    \centering
    \includegraphics[width=\linewidth]{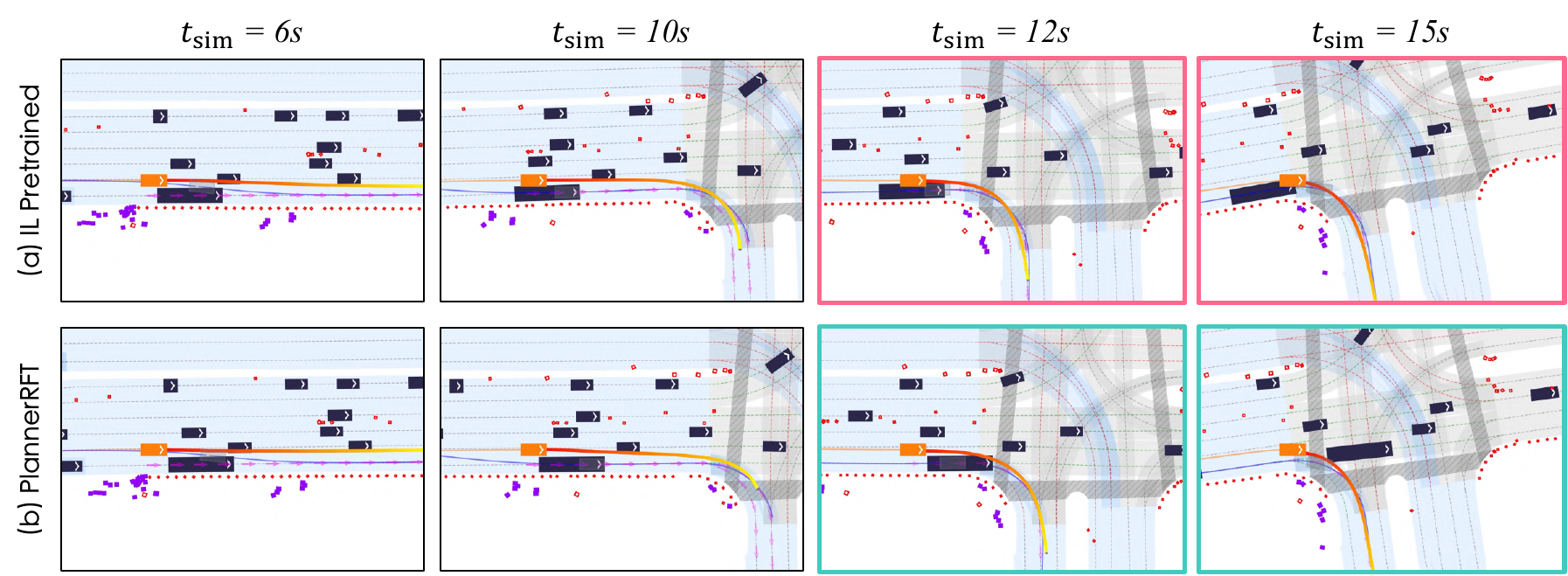}
    \caption{
        \textbf{Blocked Right-Turn in \textcolor{blue}{Reactive Traffic}.} 
        The ego vehicle intends to turn right at the upcoming intersection. 
        \textbf{(a)} The IL Pretrained planner causes the ego vehicle to forcibly change lanes, leading to a collision with a long vehicle on the right at $t_{\text{sim}}=12~s$. 
        \textbf{(b)} PlannerRFT enables the ego vehicle to consider the long vehicle proceeding straight, hence the ego vehicle decides to delay the lane change, which avoiding a collision.
    }
    \label{fig:demo_reactive1}
\end{figure*}

\begin{figure*}[t]
    \centering
    \includegraphics[width=\linewidth]{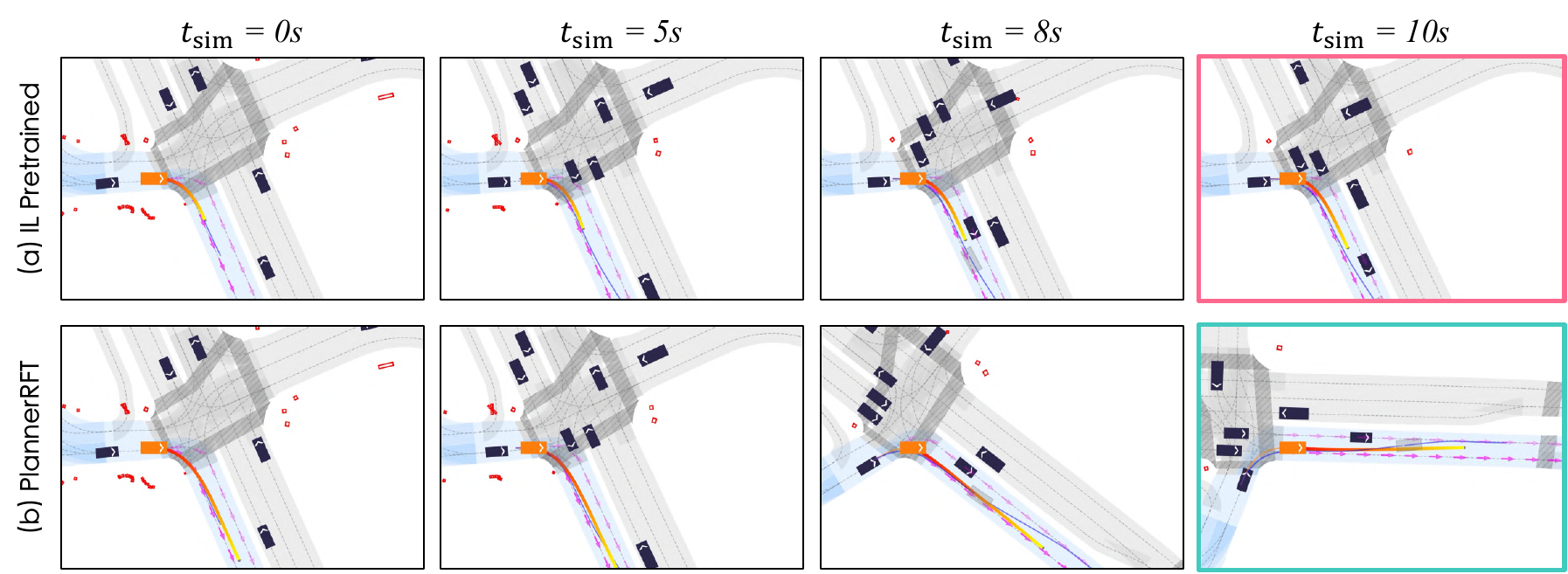}
    \caption{
        \textbf{Unprotected Right-Turn in \textcolor{blue}{Reactive Traffic}.} 
        The ego vehicle attempts a right turn at an intersection while surrounding vehicles approach from the cross traffic.
        \textbf{(a)} The IL Pretrained planner causes the ego vehicle to hesitate when turning right, ultimately leading to a collision with an oncoming vehicle from the left at $t_{\text{sim}}=10~s$. 
        \textbf{(b)} PlannerRFT enables the ego vehicle smoothly and successfully completes the right turn before the arrival of the oncoming vehicle from the left.
    }
    \label{fig:demo_reactive2}
\end{figure*}

\boldparagraph{Qualitative Results in Reactive Traffic.}
We present the closed-loop planning results for PlannerRFT in reactive traffic, as shown in \cref{fig:demo_reactive1} and \cref{fig:demo_reactive2}.
These examples demonstrate the planner’s enhanced decision-making and planning capability in interactive scenarios.

\begin{figure*}[t]
    \centering
    \includegraphics[width=\linewidth]{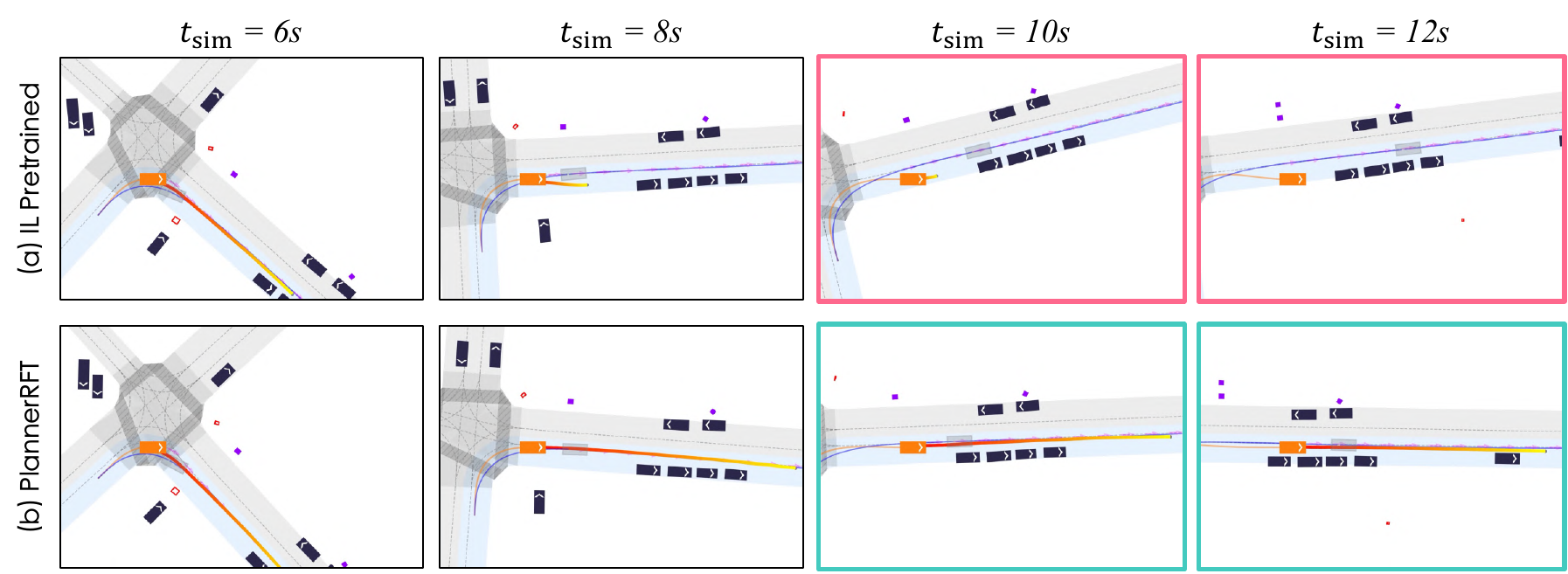}
    \caption{
        \textbf{A Causal-Confusion Scenario.} 
        The ego vehicle turns right at an intersection. 
        \textbf{(a)} The IL pretrained planner directs the ego vehicle to turn right and then pull over to the side of the road. Off-road at $t_{\text{sim}=10~s}$.
        This behavior is likely due to causal confusion: a large number of scenarios in the training data where vehicles turn right and stop to pick up passengers. 
        \textbf{(b)} PlannerRFT guides the ego vehicle to turn right and then proceed straight, a maneuver that aligns with common sense and avoids causal confusion.
    }
    \label{fig:demo_confusion}
\end{figure*}

\boldparagraph{Qualitative Results on Causal-Confusion Scenarios.}
We present closed-loop planning results on a causal-confusion scenario, as shown in \cref{fig:demo_confusion}, which illustrate the advantage of RL in mitigating the causal-confusion issues inherent in imitation learning.

\section{License of Assets}
Data for nuPlan~\cite{nuPlan} are provided under the CC-BY-NC 4.0 license.
Our IL-pretrained model follows the implementation of Diffusion Planner~\cite{DiffusionPlanner}. 
As the original repository\footnote{\url{https://github.com/ZhengYinan-AIR/Diffusion-Planner}} 
does not provide an explicit license, the referenced code is used solely for academic research 
and reproducibility purposes, and all rights remain with the original authors.
nuMax is a re-implementation of Waymax~\cite{Waymax} for non-commercial research, 
in accordance with the Waymax License Agreement for Non-Commercial Use\footnote{\url{https://github.com/waymo-research/waymax/blob/main/LICENSE}}.
Scenario caching in nuMax is developed upon ScenarioMax, which is released under the Apache-2.0 license.
Our RL training pipeline references V-Max~\cite{vmax} and Brax, 
distributed under the MIT License and Apache-2.0 License, respectively. 
The reinforcement learning algorithms further draw upon DPPO~\cite{DPPO} and ReCogDrive~\cite{ReCogDrive}, 
which are released under the MIT License and Apache-2.0 License.
All source code and models developed in this work will be made publicly available under the Apache License 2.0.

\clearpage

\end{document}